\documentclass[10pt,twocolumn,letterpaper]{article}

\usepackage{cvpr}
\usepackage{times}
\usepackage{epsfig}
\usepackage{graphicx}
\usepackage{amsmath}
\usepackage{amssymb}
\usepackage{balance}

\usepackage{bm}
\usepackage{multirow}
\usepackage{microtype}
\usepackage{listings}
\usepackage{algorithm}
\usepackage{algpseudocode}
\usepackage{mathrsfs}

\usepackage{subfigure}
\usepackage{textcomp}
\usepackage{array,multirow,graphicx}
\usepackage{float}
\usepackage[font=small,labelfont=bf,tableposition=top]{caption} 

\usepackage[pagebackref=true,breaklinks=true,letterpaper=true,colorlinks,bookmarks=false]{hyperref}

\cvprfinalcopy 

\ifcvprfinal\fi

\newcolumntype{C}{ >{\centering\arraybackslash} p{1.2cm} }
\newcolumntype{H}{ >{\centering\arraybackslash} p{2.8cm} }

\begin{document}

\def \OURS {\textit{ForensicTransfer}\xspace}
\def \FT {\textit{FT}\xspace}

\title{\vspace{-0.5cm}\OURS: Weakly-supervised Domain Adaptation for Forgery Detection}

\author{
	Davide Cozzolino\textsuperscript{1} \ \ \,\,\,
	Justus Thies\textsuperscript{2} \ \ \ \ \ \ \,\,\,
	Andreas R\"ossler\textsuperscript{2} \ \ \ \,\,\,
	Christian Riess\textsuperscript{3}\\
	Matthias Nie{\ss}ner\textsuperscript{2} \ \ \ \,\,\,
	Luisa Verdoliva\textsuperscript{1} \\
	\small \textsuperscript{1}University Federico II of Naples \ \ \ \ \ \textsuperscript{2}Technical University of Munich \ \ \ \ \ \textsuperscript{3}University of Erlangen-Nuremberg
}

\twocolumn[{%
	\renewcommand\twocolumn[1][]{#1}%
	\maketitle
	\begin{center}
	    \vspace{-0.75cm}
		\includegraphics[width=0.98\linewidth]{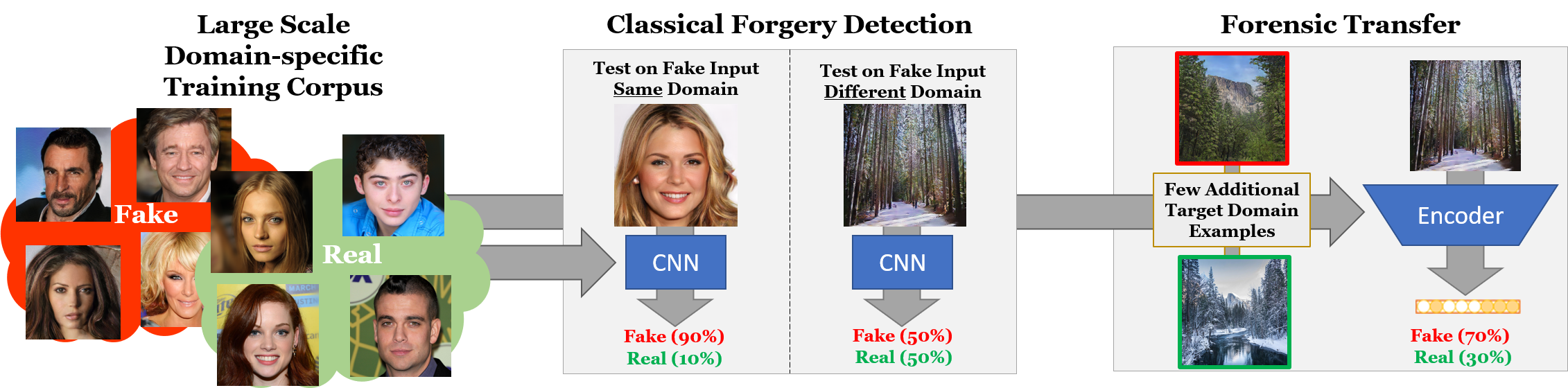}
		\vspace{-0.25cm}
		\captionof{figure}{
			CNN-based approaches for image forgery detection tend to overfit to the source training data and perform bad on new unseen manipulations.
			\OURS introduces a new autoencoder-based architecture that is able to overcome this problem given only few labeled target examples. 
			The idea is to learn a discriminative feature representation in the latent space so that only specific components of the hidden vector are activated for the pristine and forged classes.
		}
		\label{fig:teaser}
	\end{center}
}]

\begin{abstract}
\vspace{-0.25cm}
Distinguishing manipulated from real images is becoming increasingly difficult as new sophisticated image forgery approaches come out by the day.
Na\"ive classification approaches based on Convolutional Neural Networks (CNNs)
show excellent performance in detecting image manipulations when they are trained on a specific forgery method.
However, on examples from unseen manipulation approaches, their performance drops significantly.
To address this limitation in transferability, we introduce \OURS (\FT).
We devise a learning-based forensic detector which adapts well to new domains, i.e., novel manipulation methods and can handle scenarios where only a handful of fake examples are available during training.
To this end, we learn a forensic embedding based on a novel autoencoder-based architecture that can be used to distinguish between real and fake imagery.
The learned embedding acts as a form of anomaly detector; namely, an image manipulated from an unseen method will be detected as fake provided it maps sufficiently far away from the cluster of real images.
Comparing to prior works, \FT shows significant improvements in transferability, which we demonstrate in a series of experiments on cutting-edge benchmarks.
For instance, on unseen examples, we achieve up to 85\% in terms of accuracy, and with only a handful of seen examples, our performance already reaches around 95\%.
\end{abstract}

\section{Introduction}
Image manipulation is as old as photography itself \cite{Farid16}; however, with recent advances in machine learning and synthetic image rendering, it has reached unprecedented levels of diffusion and sophistication.
Manipulations that were once possible only for high-budget movie productions are now within the reach of anyone who can access large amounts of data --- a democratization of high quality image and video synthesis.
This development is the outcome of intense research in computer vision, driven by market demands in the movie industry, photography, virtual reality, game playing, and autonomous driving.
A major focus of research has been face synthesis and manipulation.
Nowadays, we can generate entirely synthetic faces \cite{Choi2018StarGAN, kingma2018Glow}, even at very high-resolutions \cite{Karras2018, Karras2019}, animate a subject's face to make it express the desired emotions \cite{elor2017bringingPortraits, Pumarola2018GANimation}, or modify facial expressions \cite{Thies2016face, suwajanakorn2017synthesizing, Kim2018deepvideo}.
Beyond faces, a number of generic image manipulation methods have been proposed.
It is now possible to transfer semantic content from a source domain to a target domain characterized by a different style \cite{Zhu2017unpaired, Karras2019}, create automatically image compositions \cite{Zhu2015learning}, and reconstruct missing image parts by semantic inpainting \cite{Liang2018generative, Iizuka2017globally}.
It is worth noting that most of these operations can be carried out using several editing tools
(e.g., \textit{Photoshop}, \textit{GIMP}), and with some manual effort achieve highly convincing results.
Although these processing tools were originally designed for benign applications, they can also be used to perform high-quality image forgeries.
Realistic fake content can have a high impact on people's lives, as demonstrated by the \textit{Deepfake app} \cite{DeepFake}.
They can be used to support fake news, and cause their viral spread over ubiquitous social networks.
Therefore, it is critical to develop methods that allow the reliable detection of such manipulations, and there has been growing attention towards this goal in the research community.
Deep neural networks have proven to be very effective for this task and several works can be found in the current literature \cite{Bayar16, Cozzolino17, Rahmouni2017, Roessler2018, Afchar2018}.
These methods heavily rely on a sufficient number of training examples and the performance impairs dramatically when new types of manipulations are presented, even though they are semantically close.
The underlying neural networks quickly overfit to manipulation-specific artifacts, thus learning features that are highly discriminatory for the given task but severely lack transferability.
This weakness can be partially addressed by fine-tuning a pre-trained network with new task-specific data, but this means that large amounts of new data are necessary.
Due to the high dynamics in the field of digital content synthesis this is not feasible and introduces a huge time delay for every new method that has to be debunked.
Ideally, one would like to detect a forgery even if the network has not been trained for it, or if only a few labeled examples are available.
The goal of this work is to develop a CNN-based method that ensures such generalization; i.e., being able to transfer knowledge between different but related manipulations.
Specifically, we suppose to have a network trained to detect a specific manipulation and want to extend its capability to similar ones.
To the best of our knowledge, this is the first work addressing this problem in the context of multimedia forensics.
To this end, we make the following contributions:
\begin{itemize}
    \item a novel autoencoder-based neural network architecture, that learns a model for a derived forensic embedding, capable of transferring between different manipulation domains,
    \item a thorough ablation study of our proposed method, including a survey on existing baseline methods,
    \item and state-of-the-art detection accuracies that enable robust forgery detection even when no or only a handful of training samples of a new manipulation are available.
\end{itemize}

\section{Related work}
While the main focus of our work lies in the field of media forensics, \OURS also intersects with the field of transfer learning.
There is a wide range of image forgery detection methods.
These methods can be divided into traditional model-based and learning-based approaches.

\paragraph{Traditional Media Forensics}

Traditional model-based approaches exploit a specific clue, like inconsistencies at pixel-level related for example to JPEG compression artifacts \cite{Agarwal2017JPEGdimples}, demosaicking clues \cite{Ferrara12:IFL}, lens aberration \cite{Yerushalmy2011digital}, or camera noise \cite{Lyu14:ERS}.
They are very effective, but also very sensitive to the assumed hypotheses.
Robustness is much more ensured by physics-based approaches that rely on inconsistencies in illumination or perspective mappings \cite{Carvalho13:EDI}; however, they are still not able to provide a performance comparable to pixel-based methods in realistic situations.

\paragraph{Learned Media Forensics}
Recently, the research community has shown that supervised deep learning approaches can achieve impressive detection results.
Similar to the traditional media forensics, the first approaches concentrated on the high-frequency pixel-level signals.
This can be accomplished by adding a first layer of fixed high-pass filters \cite{Rao16, Liu18}, learned filters \cite{Bayar16}, or even by recasting hand-crafted features working on residuals as a convolutional neural network (CNN) \cite{Cozzolino17}.
Two-stream networks are used to exploit both low-level and high-level features \cite{Zhou18}.
Other architectures try to exploit sharp discontinuities created by the editing process, such as the boundary artifacts occurring when an object is spliced into the image  \cite{Salloum18, Bappy2017}.
A significant performance gap can be observed by very deep networks, especially, on compressed data \cite{Roessler2018}, which is critical to detect fake content on social networks.
All proposed learning-based methods need some form of finetuning on a dataset that comprises manipulations aligned with the ones present in the test set.
But often the underlying datasets are limited and prone to polarization, e.g., when a limited number of cameras is used.
To avoid any form of polarization on the training-set, recent approaches have adopted Siamese networks to train couples of patches \cite{Huh2018, Cozzolino2020}.
This way, they do not need to rely on a specific training set, since they are acting as anomaly detectors on a pixel-level basis. 
Hence, their performance is much more effective on localization than on detection.
Recently, there are has been some effort in the detection of  
more recent manipulations that have been created using deep learning (i.e. deepfakes).
Some papers have focused their effort in the detection 
of manipulations on faces \cite{Afchar2018, Guera18deepfake} or GAN generated images \cite{Marra2018, Zhang2019}.
In contrast to these methods, we are not restricted to a specific type of manipulation and only need a few samples to adapt to new manipulations.

\paragraph{Transfer learning}
Transfer learning is an important problem in the vision community, especially, in deep learning which relies on a large amount of training data compared to traditional machine learning approaches.
When the distributions of training and test data do not match (domain shift), the network has to be re-trained on a large labeled dataset, and this is not easily feasible.
In the literature several solutions have been proposed to adapt the source domain to the target domain in different scenarios \cite{Csurka2017,hoffman2017cycada,tzeng2017adversarial}.
A possible direction is to use an autoencoder to learn a latent embedding in the source domain from which the feature space in the target domain is derived \cite{Glorot2011, Chen2012}.
This is as also carried out in \cite{Kodirov2017} to learn a better semantic representation in the \textit{Zero Shot Learning} scenario.
The \textit{Few Shot Learning} scenario is particularly interesting in our context, since it helps learning generalization using a limited amount of labeled examples.
\OURS is designed for the forensic analysis of imagery, i.e., we are interested in two classes -- pristine vs forged images.
This restriction is different to the traditional setting in the computer vision community which focuses on a large number of classes, resulting in approaches that aim at learning good model initialization, learning image similarity or learning to augment data through a proper generator~\cite{Chen2019}.

\section{Proposed Method}
\label{sec:method}

We propose a new CNN-based image manipulation detector which provides state-of-the-art performance on known manipulations {\em and} generalizes well to new forms of manipulations.
\OURS:
{\it   i)} provides the same excellent performance as other deep nets over known manipulations, when plenty of labeled examples are available,
{\it  ii)} generalizes better than these competitors to unseen manipulations (zero-shot performance), and
{\it iii)} improves effectively its performance for new manipulations based on a very small number of training examples (few-shot performance).

Our approach disentangles the information necessary to make the real/fake decision in the source domain from a faithful latent-space representation of the image, which may be exploited in new target domains.
To prevent the net from discarding precious information during training, we rely on autoencoder-based representation learning \cite{Tschannen2018} by which the latent space is constrained to preserve all the data necessary to reconstruct the image in compact form.

Therefore, the latent space holds both the image representation and the data used for the real/fake decision, but these pieces of information live in orthogonal spaces, and do not interfere with one another.
This is obtained by dividing the latent space in two parts, one activated exclusively by real samples, and the other by fake samples.
Since the network has to reproduce the image anyway, all relevant information on the input image is stored in both parts.
Thus, the features of the learned forensic embedding keep all desired information, useful for diverse forensic tasks, and easily adapted to new domains based on a small number of new training samples.

\paragraph{Forensic Embedding:}
Initially, we assume to have only training data for the source domain, formed by
the set $\mathcal{S}_{1}=\left\{x^{s1}_i \right\}^N_{i=1}$ comprising images tampered by the known manipulation, and the set $\mathcal{S}_{0}=\left\{x^{s0}_i \right\}^N_{i=1}$ of the corresponding original images (or real images of the same typology if the original images do not exist).
Target domain manipulations are defined correspondingly as $\mathcal{T}_{1}$ and $\mathcal{T}_{0}$.

As a preprocess, the input images are high-pass filtered, in order to obtain the residual image which is known from the multimedia forensics literature to store the most valuable information for forgery detection.
In particular, we apply a third-order derivative in image space~\cite{Cozzolino17}.

Our proposed detector is a neural network with an autoencoder structure where
the encoding function ${\cal E}(\cdot)$ maps the image $x$ to the latent vector $h$, and
the decoding function ${\cal D}(\cdot)$ processes the latter to provide an approximate reconstruction $\hat{x}$, of the input image.
To disentangle decision and reconstruction,
the latent vector $h$ is split in two disjoint parts, $h_0$ and $h_1$, each associated with a class, real or fake.
Accordingly,
the network is trained to activate only the part $h_0$ if $x$ belongs to $\mathcal{S}_{0}$ and only $h_1$ if $x$ belongs to $\mathcal{S}_{1}$.
At testing time, the input image is classified as forged or real by measuring the activations of the two parts.
\begin{figure}
	\centering
	\includegraphics[width=0.9\linewidth, trim=0 130 190 0,clip]{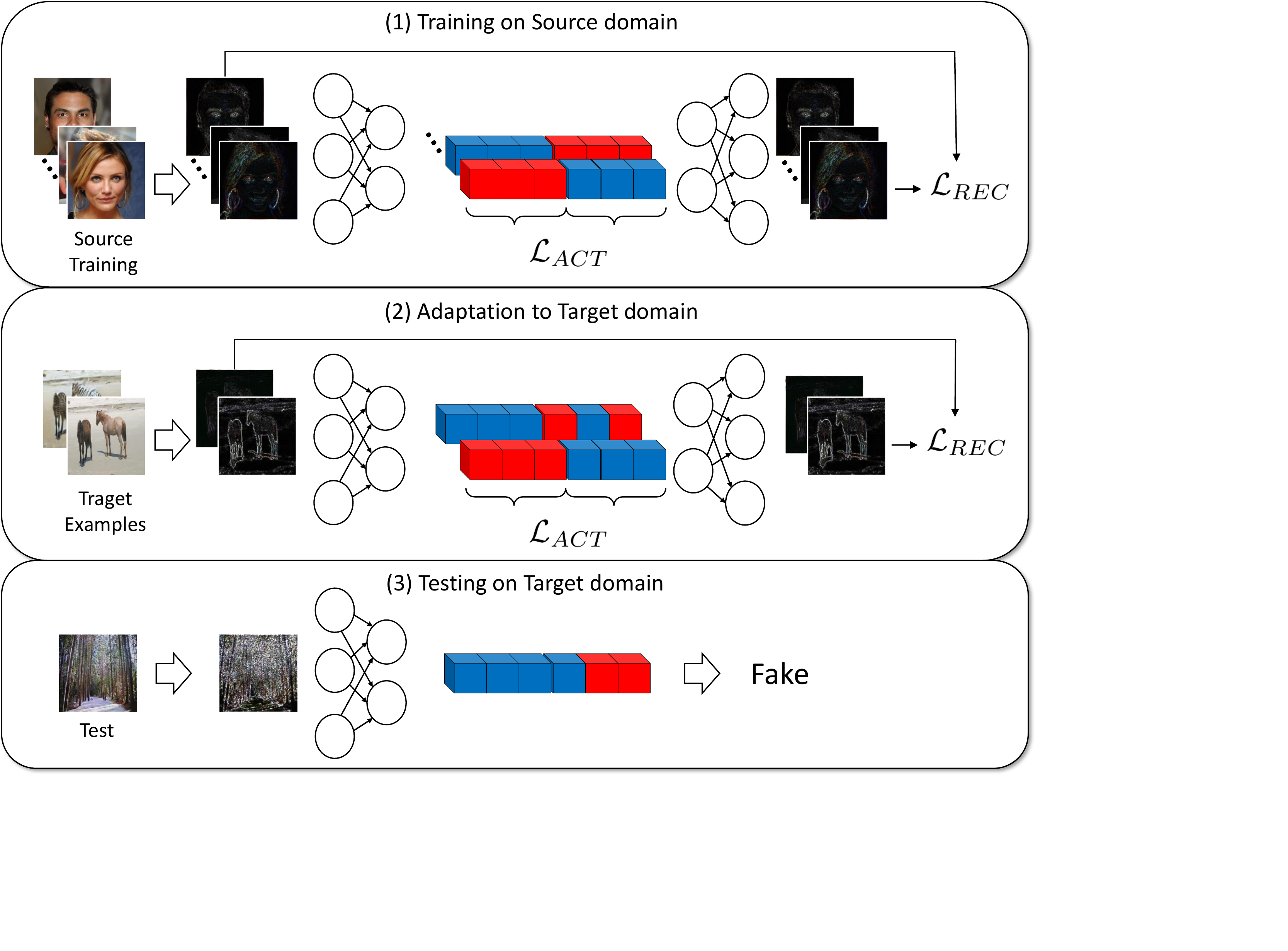}
	\vspace{-0.15cm}
    \caption{\OURS scheme. (1) We train our autoencoder-based approach on a source domain, disentangling the fake and real images in the latent space. (2) Using few target training samples one can adapt to the target domain. (3) The latent space is used to predict the class.}
    \label{fig:schema}
    \vspace{-0.4cm}
\end{figure}
The class activation $a_c(x)$ is measured through the $\ell_1$-norm of the corresponding part of the latent vector, that is $a_c(x) = \frac{1}{K_c} \left\| {\cal E}_c(x) \right\|_{\scriptscriptstyle 1}$,
where $K_c$ is the number of elements in $h_c$ of class $c \in \{0,1\}$.

For a sample $x$ of class $c$, the off-class part of the latent space $h_{1-c}$ should remain silent, {\it i.e.}, $a_{1-c}(x)=0$.
On the contrary, the in-class part of the latent space $h_{c}$ should be active, with at least one non-zero element, such that $a_{c}(x)>0$.
For simplicity, and without losing generality, we constrain $a_{c}(x)$ to be equal to $1$.

\paragraph{Loss:}
During the training phase, we enforce a reconstruction loss $\mathcal{L}_{REC}$ on the output image of the autoencoder and an activation loss $\mathcal{L}_{ACT}$ on the latent space:
\begin{equation*}
    \mathcal{L} = \gamma \cdot \mathcal{L}_{REC} + \mathcal{L}_{ACT} \enspace{,}
    \label{equ:loss_tot}
\end{equation*}
where $\gamma$ weights the influence of the reconstruction error, and is set to $0.1$ for all our experiments.
The reconstruction loss $\mathcal{L}_{REC}$ measures the difference of the input image $x$ and the reconstructed image $\hat{x}$ which is given by ${\cal D}({\cal E}(x))$ using an $\ell_1$-norm:
\begin{equation*}
    \mathcal{L}_{REC} = \frac{1}{K} \cdot \sum_{x \in \mathcal{S}_0 \cup \mathcal{S}_1} \left\| x-\hat{x} \right\|_{\scriptscriptstyle 1} \enspace{,}
\end{equation*}
where $K$ is the number of components of the input sample.
The reconstruction constraint forces the net to store all relevant data of the input in the latent vector.
Hence, we avoid to encode only the information directly useful for discriminating the source manipulation, allowing adaptation to novel manipulations.
The activation loss $\mathcal{L}_{ACT}$ is defined as:
\begin{eqnarray*}
    \mathcal{L}_{ACT} & = & \sum_{x \in \mathcal{S}_0} \left| a_0(x)-1 \right| + \left|a_1(x) \right| ~+ \nonumber \\
                      &  & \sum_{x \in \mathcal{S}_1} \left| a_1(x)-1 \right| + \left|a_0(x) \right|
    \label{equ:loss_L1_act}
\end{eqnarray*}
At test time, the decision on the presence of a forgery relies on the strength of the activations.
A sample $x$ is considered to be forged ($c=1$) if $a_1(x)>a_0(x)$, and real otherwise.
Unlike the widespread classification loss based on cross-entropy, the proposed loss does not only force the separation of the two classes, but also the reduction of the intra-class variances, which improves transferability in our forensic application.
As a consequence, an unseen novel manipulation can be more easily distinguished from the class of real images.
In the literature, the importance of reducing intra-class variances has been demonstrated in \cite{Hu2015} for transfer learning and in \cite{Chen2019} for few-shot problems.

\paragraph{Transferability:}
Based on strength of the activations, we are able to decide whether the input image is closer to the real or to the forged images of the source domain.
Having to deal with a novel manipulation, where only a few examples are available, we exploit a detector already trained on a ``close'' manipulation, which can be assumed to share some artifacts with the new one.
Therefore, we fine-tune this detector to the target domain using the available examples in $\mathcal{T}_{1}$ and $\mathcal{T}_{0}$ and the loss $\mathcal{L}$.

\begin{figure}
    \centering
    \includegraphics[width=1.0\linewidth, clip, page=2, trim=0 0 330 0]{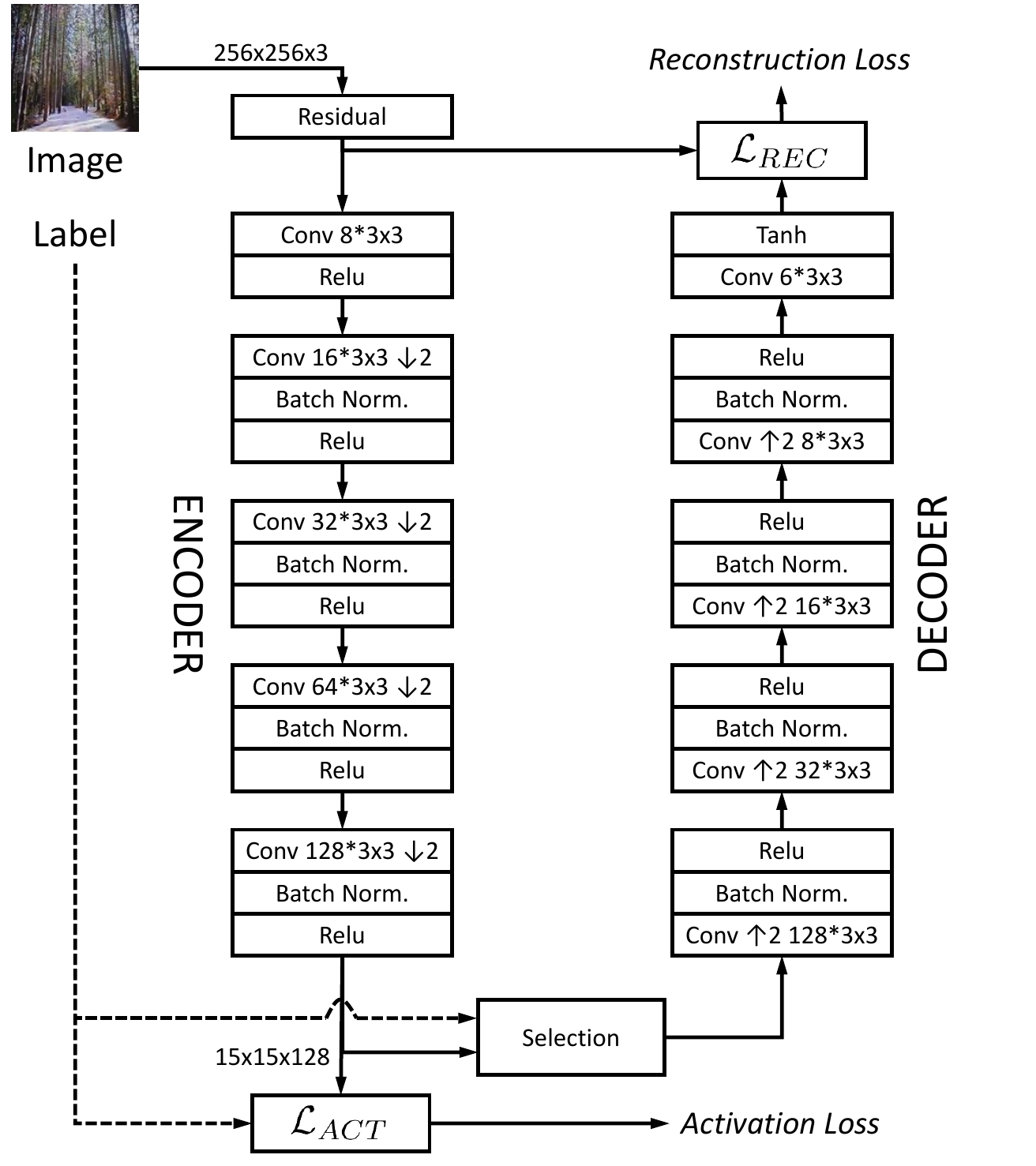}
    \vspace{-0.25cm}
    \caption{\OURS neural network architecture. As input we take an image from which we first derive the residual image, which is then fed into an encoder (left). The learned embedding is constrained by the activations of the class loss (fake vs real), as well as the reconstruction loss of the decoder (right).}
    \vspace{-0.35cm}
    \label{fig:architecture}
\end{figure}

\paragraph{Network architecture}
In this work we consider the encoder-decoder architecture shown in Fig.\ref{fig:architecture}.
Input images have size $256 \times 256 \times 3$ or $128 \times 128 \times 3$ pixels, depending on the dataset (see Sec.~\ref{sec:results}).
The encoder and decoder have mirrored structures, with $5$ convolutional layers with a $3 \times 3$ kernel each, and without skip connections.
In the encoder, all convolutions, except the first one, have stride $2$,
thus reducing spatial resolution by a factor $16$ overall.
To recover the original size, the decoder employs a $2 \times 2$ nearest-neighbor up-sampling before each convolution except the last one.
All activation functions are ReLUs (Rectified Linear Units)
except for the hyperbolic tangent activation used in the last layer to ensure the output to be in a pre-defined limited range.
The latent space at the encoder output, comprises $128$ feature maps, $64$ associated with the class ``real'' and $64$ with the class ``fake''.
The selection block sets the off-class part of latent space to zero, depending on the class label, forcing the decoder to learn how to reconstruct the input sample only from the same-class part of the latent space.

\paragraph{Training}

To evaluate its generalization capability, the network is first trained on a dataset built w.r.t. a specific manipulation method (source domain).
Then, it is fine-tuned on a few images subject to a different, but similar, form of manipulation (target domain).
The fine-tuning procedure allows the detection network to quickly adapt to the new manipulation scenario, without the need to be re-trained.

For training, we use the ADAM gradient-based optimization scheme with a learning rate of $0.001$, a batch-size of $64$ and default values for the moments ($\beta_1=0.9$, $\beta_2=0.999$ and epsilon$=10^{-7}$).
We train the network until the loss, evaluated on the validation, does not improve for $30$ consecutive epochs.
\section{Datasets}
\label{sec:datasets}

\begin{figure}[t!]
	\centering
	\includegraphics[width=0.98\linewidth,trim=0 20 0 20, page=1]{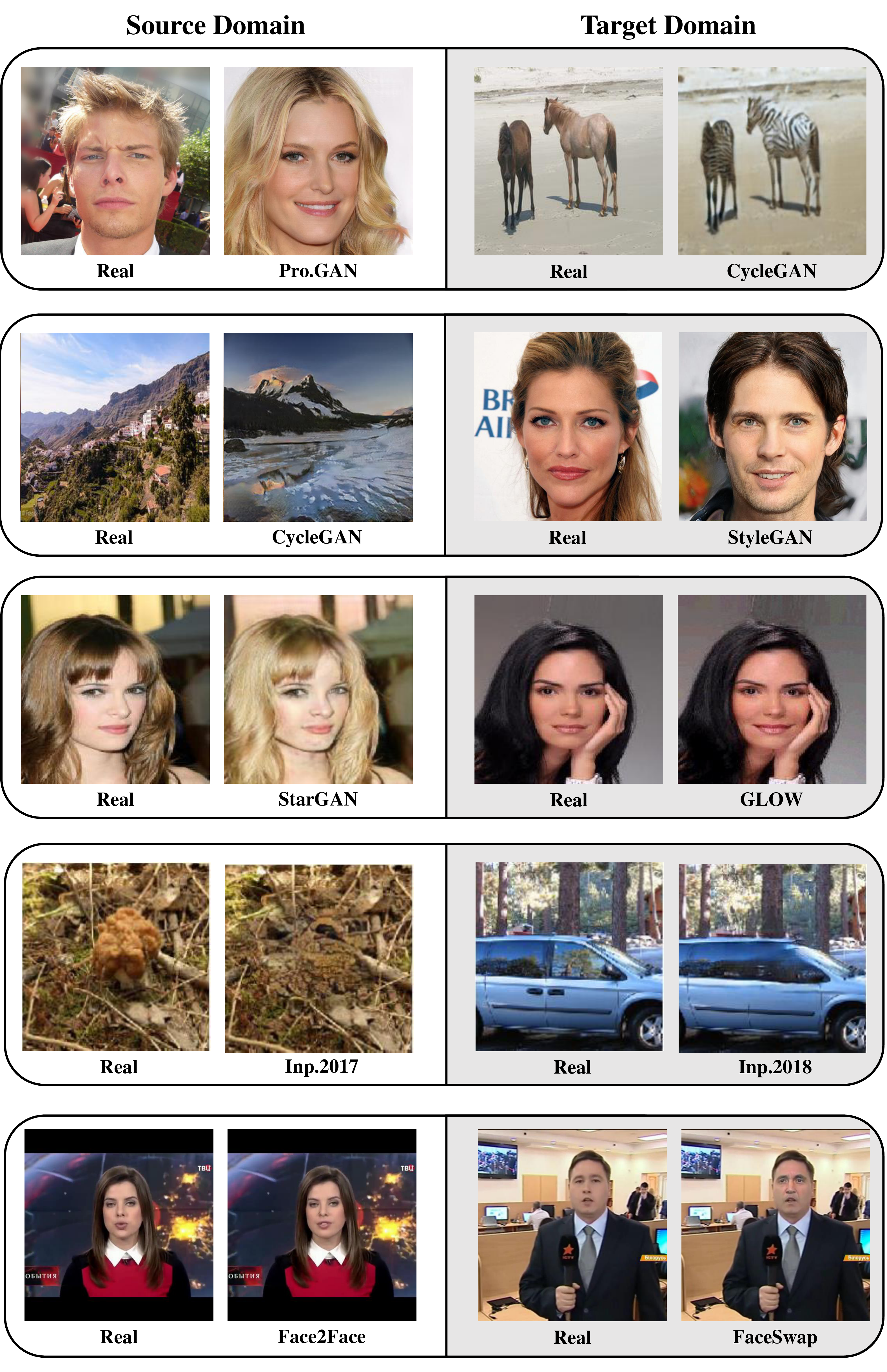}
	\caption{Examples of the different datasets. Source domain (left), target domain (right).    }
	\label{fig:datasets}
\end{figure}

In our experiments we focus on manipulations performed by computer graphics (CG) or deep learning methods,
which may involve parts of real images (for example, changes only in the mouth) 
or the generation of whole synthetic images (see Fig.~\ref{fig:datasets}).
In all cases, we used large datasets for the source domain and
split each dataset in training, validation and test.
The few examples of the target dataset, used for fine-tuning, are randomly extracted from the training and validation sets.
Note that we took care to pair two different but related manipulations in all our experiments for the source and target dataset.
In the following we describe each dataset.
For easy reproducibility of our results, we will publish our new datasets of synthetic and inpainted images.

\paragraph{Synthetic images:}
Because of the lack of publicly available dataset for forensic research purposes, we built $5$ datasets of $30000$ images each comprising images generated using progressive GAN \cite{Karras2018}, Cycle-GAN \cite{Zhu2017unpaired}, Style-GAN \cite{Karras2019}, Glow \cite{kingma2018Glow}, and StarGAN \cite{Choi2018StarGAN}.
Progressive GAN and Style-GAN generate high resolution faces, Cycle-GAN performs image-to-image translation,
and the last two methods change attributes of a face.
Except for the high resolution images of dimension $1024 \times 1024$, all the other output images are $256 \times 256$ pixels.
The total number of $30000$ images per manipulation method is split into $21000$, $6000$ and $3000$ image set for training, validation and test, respectively.
For our test we used the progressive GAN, Cycle-GAN and StarGAN images as the source datasets while Cycle-GAN, Style-GAN and Glow are chosen to be the target datasets, respectively.
Real faces come from CelebA-HQ dataset \cite{Karras2018} for ProGAN and StyleGAN,
CelebA dataset \cite{Liu2015celabA} for StarGAN and Glow, 
and from \cite{Zhu2017unpaired} for Cycle-GAN.

\paragraph{Inpainted images:}

We built another two datasets of $20000$ images each using inpainting-based manipulation methods.
To this end, we created a source database using the method of Iizuka et al.~\cite{Iizuka2017globally} and a target database using Yu et al.~\cite{Yu2018generative}.
The inpainting is applied to the central $128 \times 128$-pixel region of the image.
Since only the central region is manipulated, we considered only this part as the input for the networks.
The set of  $20000$ images are split into $14000$, $3000$, $3000$ images for training, validation and test, respectively.
Original images come from ImageNet \cite{ILSVRC15}.

\paragraph{CG-based manipulated faces:}

We are using the public dataset \textit{FaceForensics}, proposed in \cite{Roessler2018}, with $1004$ real videos and $1004$ fake ones manipulated by \textit{Face2Face} \cite{Thies2016face}.
We used the same original videos to create another dataset of manipulated videos using \textit{FaceSwap} \cite{FaceSwap}.
The \textit{Face2Face} generated images are used as the source dataset and the \textit{FaceSwap} manipulation as the target dataset.
The $1004$ videos are divided into $704$ for training, $150$ for validation and $150$ for testing.
All videos have been compressed using H.264 with quantization parameter set to $23$.
During the training we used $200$ frames from each video, while the testing is performed on ten frames per video.
For each frame, we cropped all images to be centered around the face.

\begin{figure*}[t!]
	\begin{center}
	{\renewcommand{\arraystretch}{0.0}
		\begin{tabular}{cc}
			\multicolumn{2}{c}{\includegraphics[width=0.6\linewidth, page=2]{figures/preliminary_inp2.pdf}}  \\
			\hspace{-0.45cm}
			\includegraphics[width=0.485\linewidth, page=1, trim=0 -5 0 0]{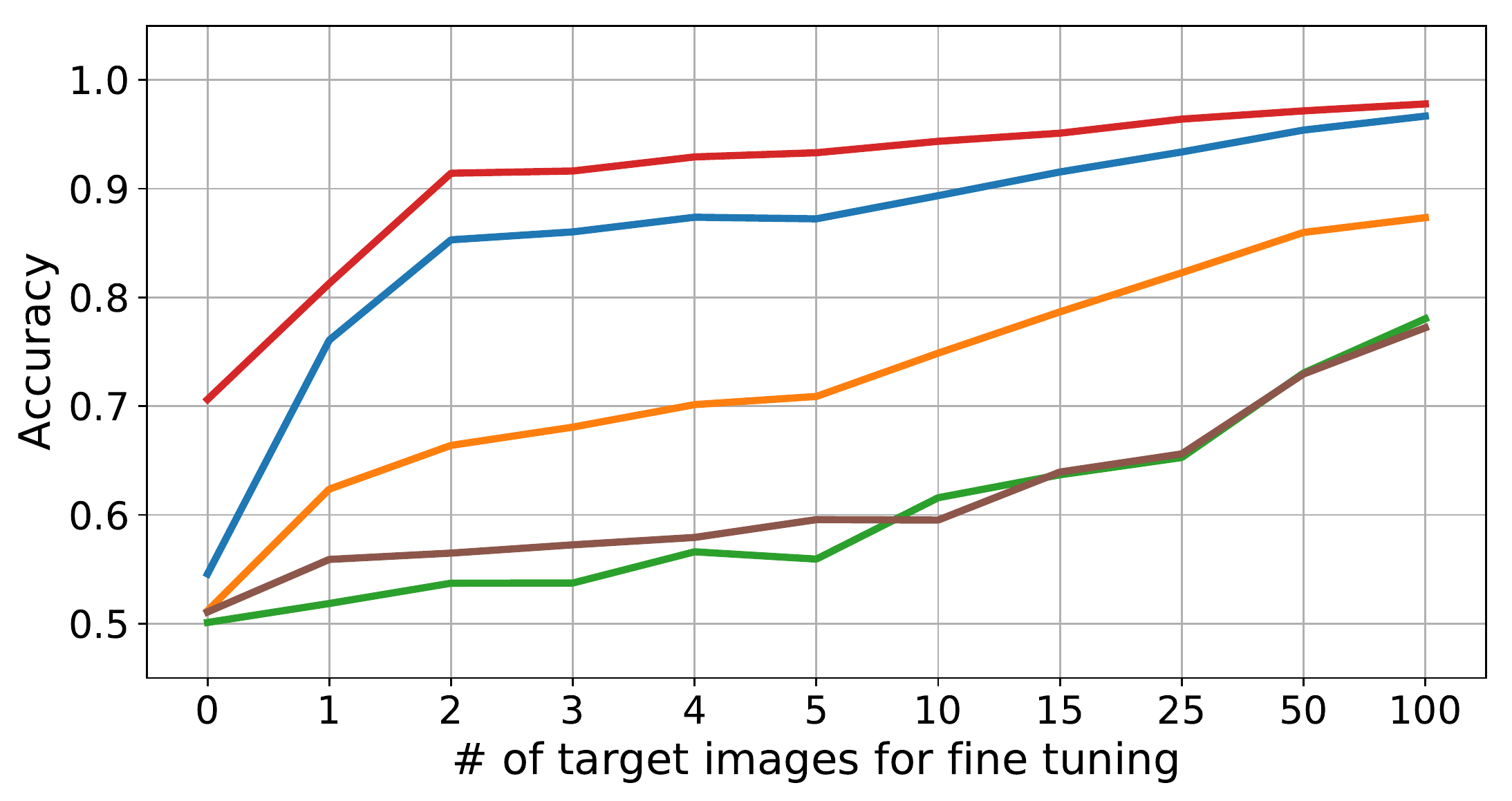} &
			\includegraphics[width=0.485\linewidth, page=1, trim=0 -5 0 0]{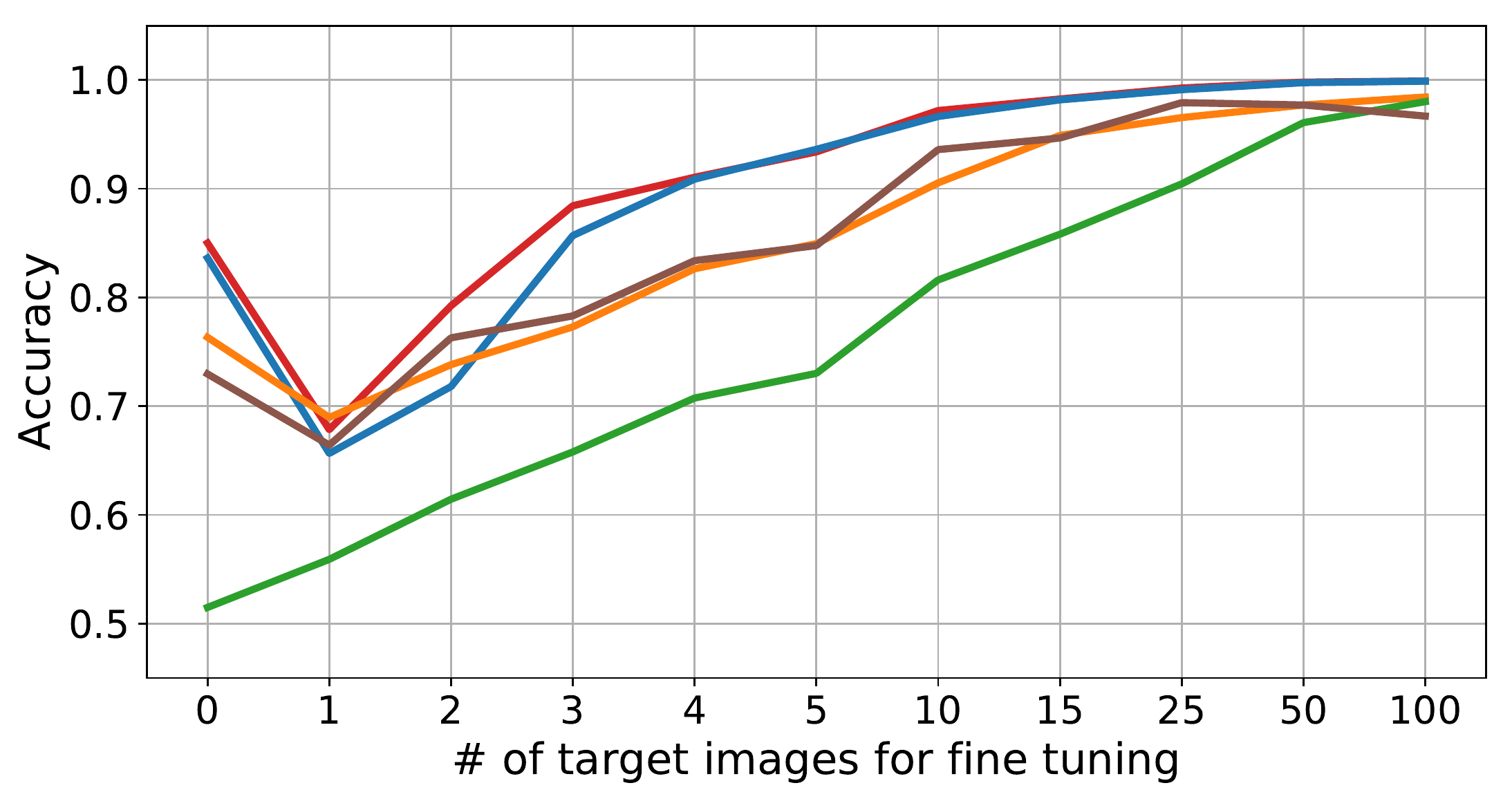}  \\
			{\footnotesize (a) Inpainting \cite{Iizuka2017globally} vs \cite{Yu2018generative} } &
			{\footnotesize (b) Pro.GAN \cite{Karras2018}  vs Cycle \cite{Zhu2017unpaired} } \\
		\end{tabular}
	}
	\end{center}
	\vspace{-0.5cm}
	\caption{
		Comparison of the proposed method with three variants.
		Plots show the accuracy, averaged over 10 runs, versus number of images used for fine-tuning.
	}
	\label{fig:preliminary_exp}
	\vspace{-0.25cm}
\end{figure*}

\begin{figure*}[t!]
\begin{center}
\includegraphics[width=\linewidth, page=1, trim=0 0 0 0]{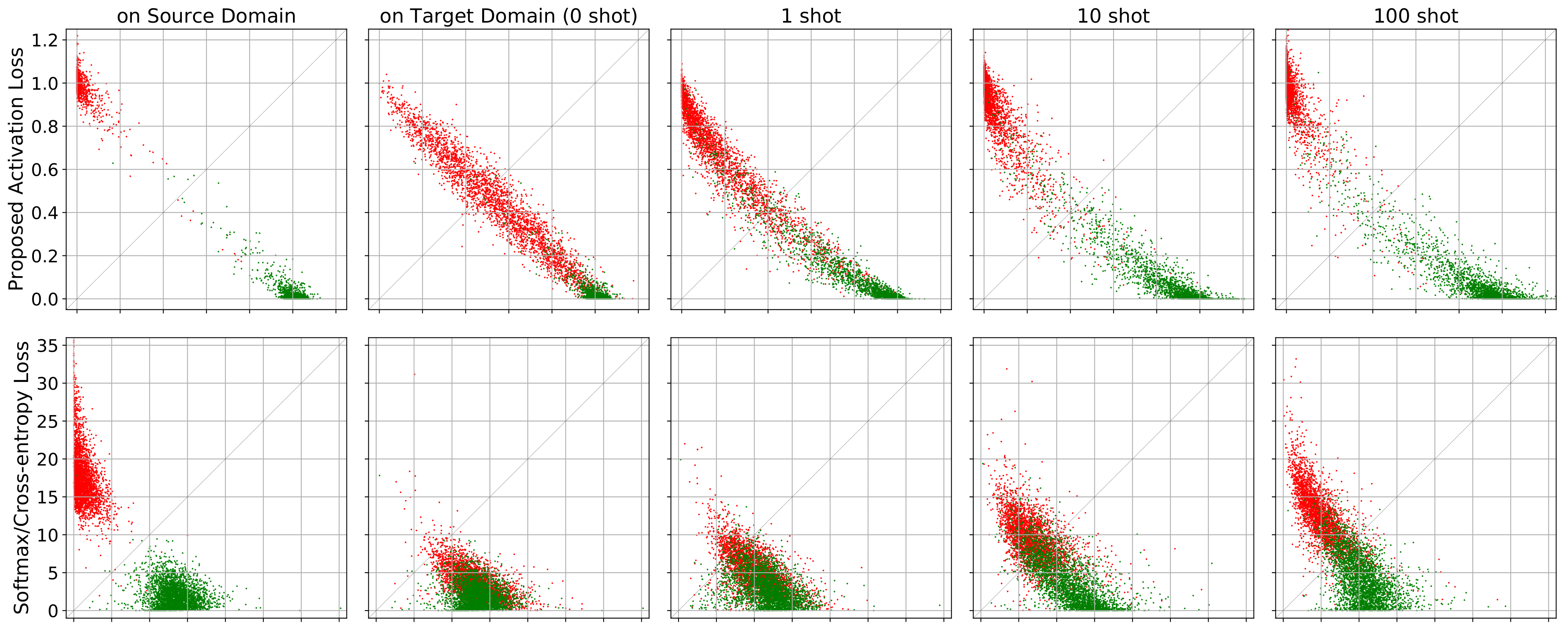}
\end{center}
\vspace{-0.65cm}
   \caption{ 
   Scatter plots of the two activations for images manipulated with inpainting, \cite{Iizuka2017globally} vs \cite{Yu2018generative}. 
   The two rows show scatter plots related to our proposal as well as results where the proposed activation function is replaced by the softmax/cross-entropy loss.
   }   \vspace{-0.4cm}
\label{fig:preliminary_exp_scatter}
\end{figure*}

\section{Results}
\label{sec:results}
In this section we give a thorough analysis of our approach.
We show the impact of the design choices (see Sec.~\ref{sec:preliminary}),
and show state-of-the-art detection results in comparison to forensic approaches and few-shot learning methods (see Sec.~\ref{sec:F_comparison} and supplemental document).
We also analyze the influence of having multiple source domains on the detectability of fakes in a target domain (see Sec.~\ref{sec:muli_source}).

\begin{table*}
    \begin{minipage}{0.67\textwidth}
        \centering
        \newcommand{\ru}{\rule{0mm}{3mm}}
        \renewcommand{\tabcolsep}{5pt}
        {\footnotesize
        \centering
        \begin{tabular}{l||c|c|c|c|c|c|c|c|c|c|} \cline{2-11}
        \ru             & \multicolumn{2}{c|}{Pro/Cycle} & \multicolumn{2}{c|}{Cycle/Style} & \multicolumn{2}{c|}{Star/Glow} & \multicolumn{2}{c|}{Inpainting} & \multicolumn{2}{c|}{Face2/Swap} \\ \cline{2-11}
        \ru             & S & T & S & T & S & T & S & T & S & T \\ \hline\hline
        \ru Bayar       & 99.92 & 50.42 & 98.92 & 55.93 & 99.98 & 50.00 & 99.08 & 50.10 & 92.93 & 52.87 \\ \hline
        \ru Cozzolino   & 99.92 & 52.43 & 99.92 & 49.98 & 100.0 & 50.00 & 98.12 & 63.33 & 79.80 & \textbf{77.77} \\ \hline
        \ru Rahmouni    & 100.0 & 49.87 & 98.85 & 62.52 & 100.0 & 50.00 & 95.35 & 52.02 & 93.57 & 70.87 \\ \hline
        \ru MesoInc.    & 97.47 & 44.19 & 87.98 & 50.30 & 99.50 & 49.97 & 87.65 & 67.30 & 85.87 & 45.17 \\ \hline
        \ru Xception    & 100.0 & 58.79 & 99.92 & 50.08 & 100.0 & 50.00 & \textbf{99.98} & 51.08 & \textbf{98.13} & 50.20 \\ \hline \hline
        \ru FT (ours)   & \textbf{100.0} & \textbf{85.00} & \textbf{100.00} & \textbf{90.53} & \textbf{100.0} & \textbf{82.05} & 99.77 & \textbf{70.62} & 94.47 & 72.57 \\ \hline
        \end{tabular}
        \vspace{0.25cm}
        \caption{Accuracy on the Source dataset and Target dataset. All the methods perform very well on the manipulations seen in the training set and much worse on new ones.}
        \label{tab:my_label}
        \vspace{-0.45cm}
        }
    \end{minipage}
    \hfill
    \begin{minipage}{0.3\textwidth}
         \centering
         \newcommand{\ru}{\rule{0mm}{3mm}}
         \renewcommand{\tabcolsep}{5pt}
        {\footnotesize
        \begin{tabular}{|l||c|} 
    \multicolumn{2}{c}{Accuracy on CycleGAN}   \\ \hline \hline
    \ru  one-to-one~~~~(Pro.GAN)     & 85.00  \\ \hline
    \ru  one-to-one~~~~(StarGAN)     & 77.26  \\ \hline
    \ru  one-to-one~~~~(Glow)        & 83.67  \\ \hline\hline
    \ru  multi-to-one~(StarGAN+Glow) & \bf{92.32}  \\ \hline
        \end{tabular}
        \vspace{0.2cm}
        \caption{
            Our approach enables the transfer of detection abilities from multiple source domains to a target domain.
            We show zero-shot results using our approach based on a one-to-one transfer in comparison to a multi-to-one transfer.
        }
        \vspace{-0.2cm}
        \label{tab:results_multi}
        }
    \end{minipage}
\end{table*}

\begin{figure*}
    \centering
    {\renewcommand{\arraystretch}{0.0}
    \begin{tabular}{cc}
       \multicolumn{2}{c}{\includegraphics[width=0.9\linewidth, trim=0 10 0 0]{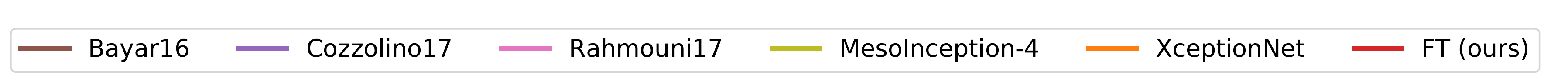}}  \\
       \includegraphics[width=0.485\linewidth, page=1, trim=0 -5 0 -10]{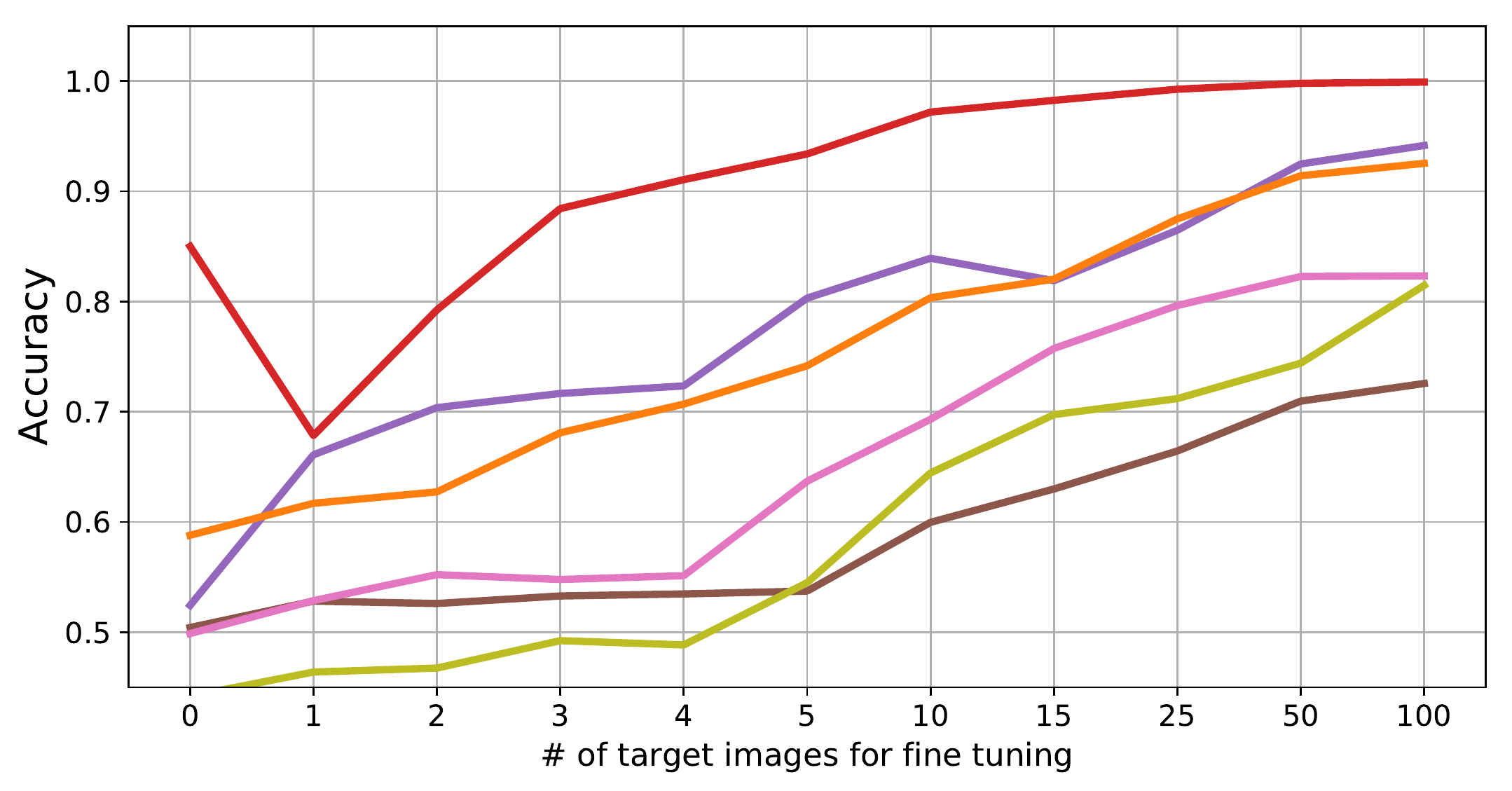} &
       \includegraphics[width=0.485\linewidth, page=1, trim=0 -5 0 -10]{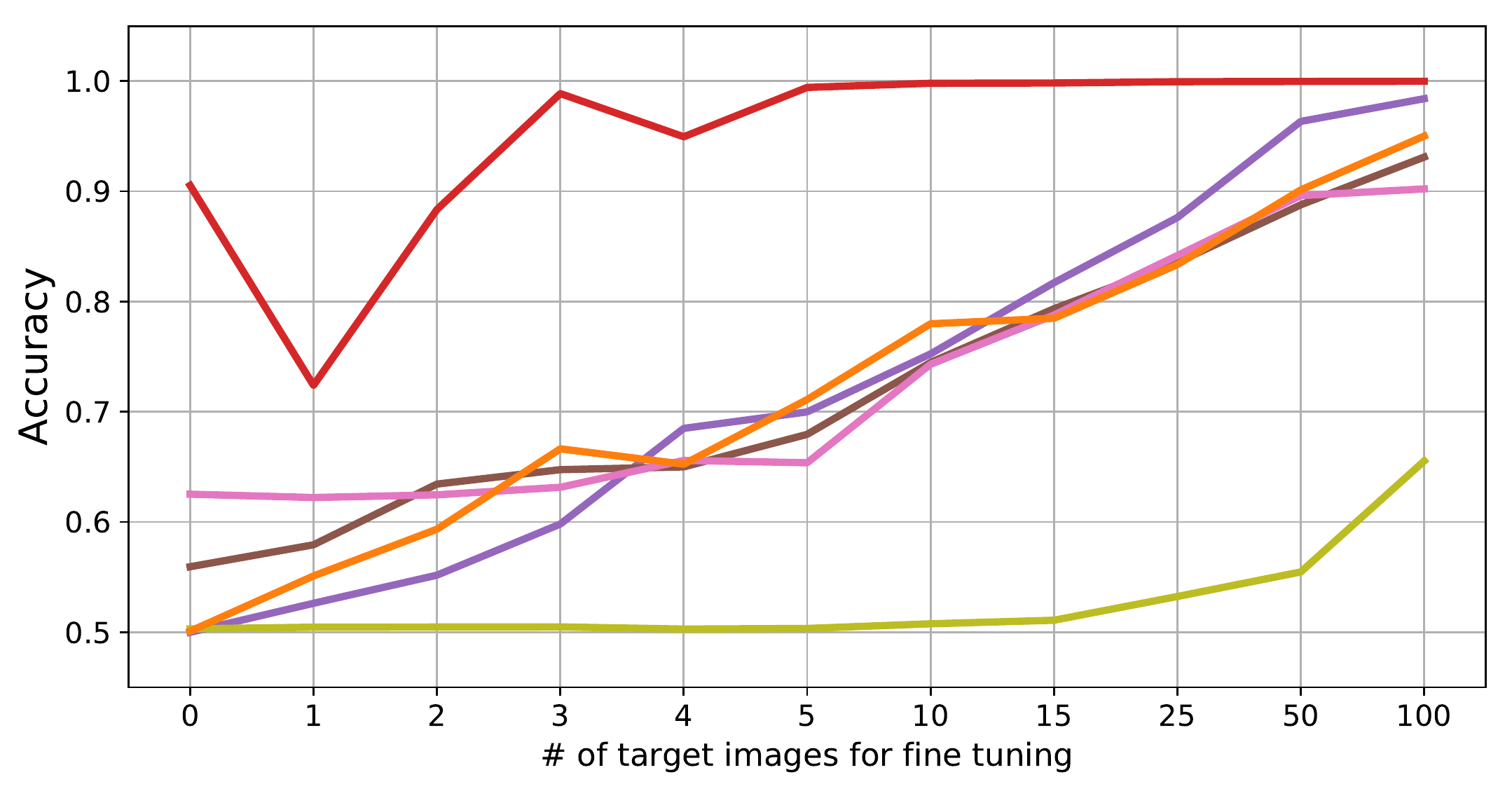} \\
       {\footnotesize (a) Pro.GAN \cite{Karras2018}  vs Cycle \cite{Zhu2017unpaired} } &
       {\footnotesize (b) Cycle \cite{Zhu2017unpaired} vs Style-GAN \cite{Karras2019} } \\
       \includegraphics[width=0.485\linewidth, page=1, trim=0 -5 0 -10]{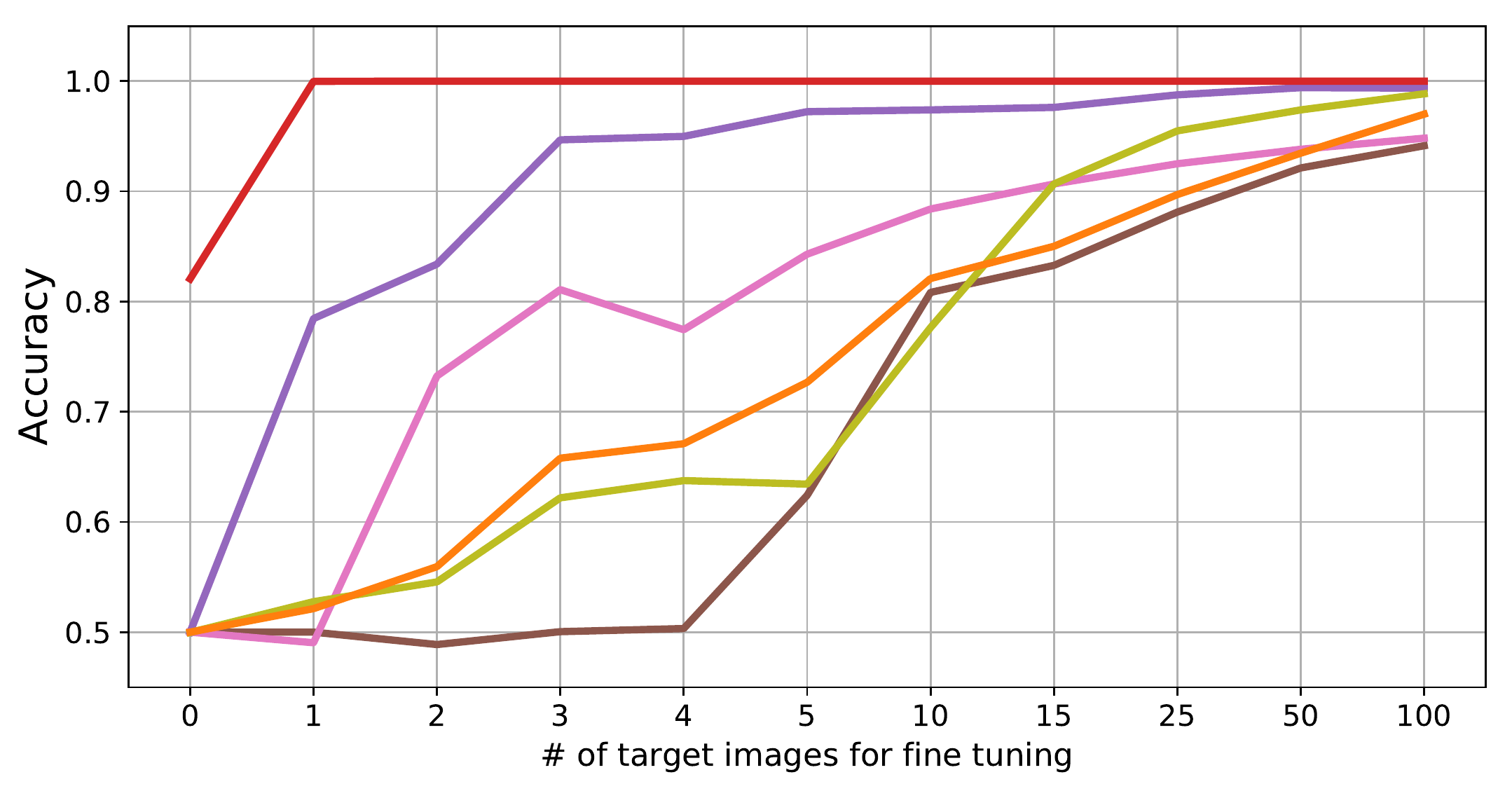} &
       \includegraphics[width=0.485\linewidth, page=1, trim=0 -5 0 -10]{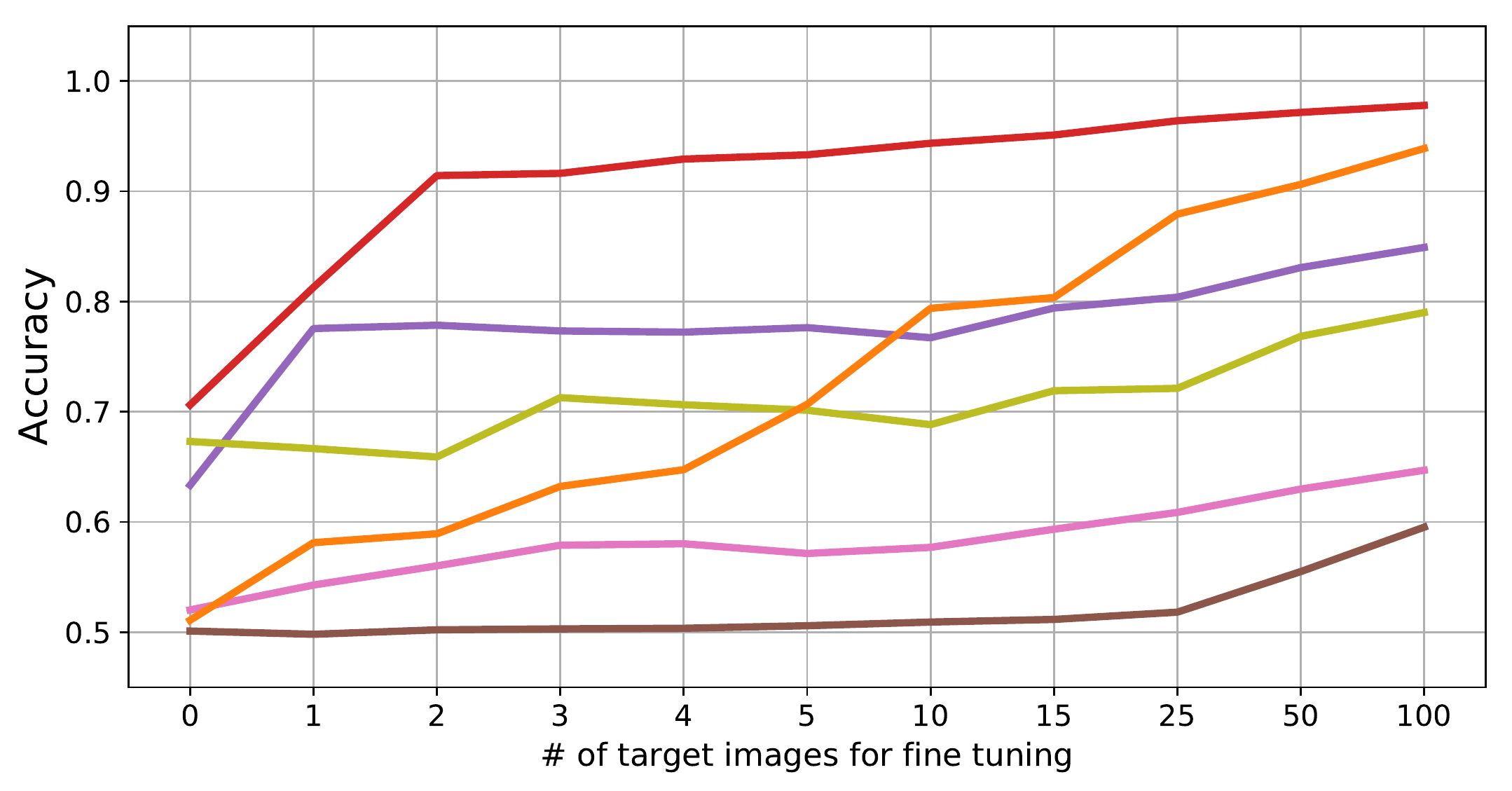} \\
       {\footnotesize (c) StarGAN \cite{Choi2018StarGAN}  vs Glow \cite{kingma2018Glow} } &
       {\footnotesize (d) Inpainting \cite{Iizuka2017globally} vs \cite{Yu2018generative}  }  \\
       \includegraphics[width=0.485\linewidth, page=1, trim=0 -5 0 -10]{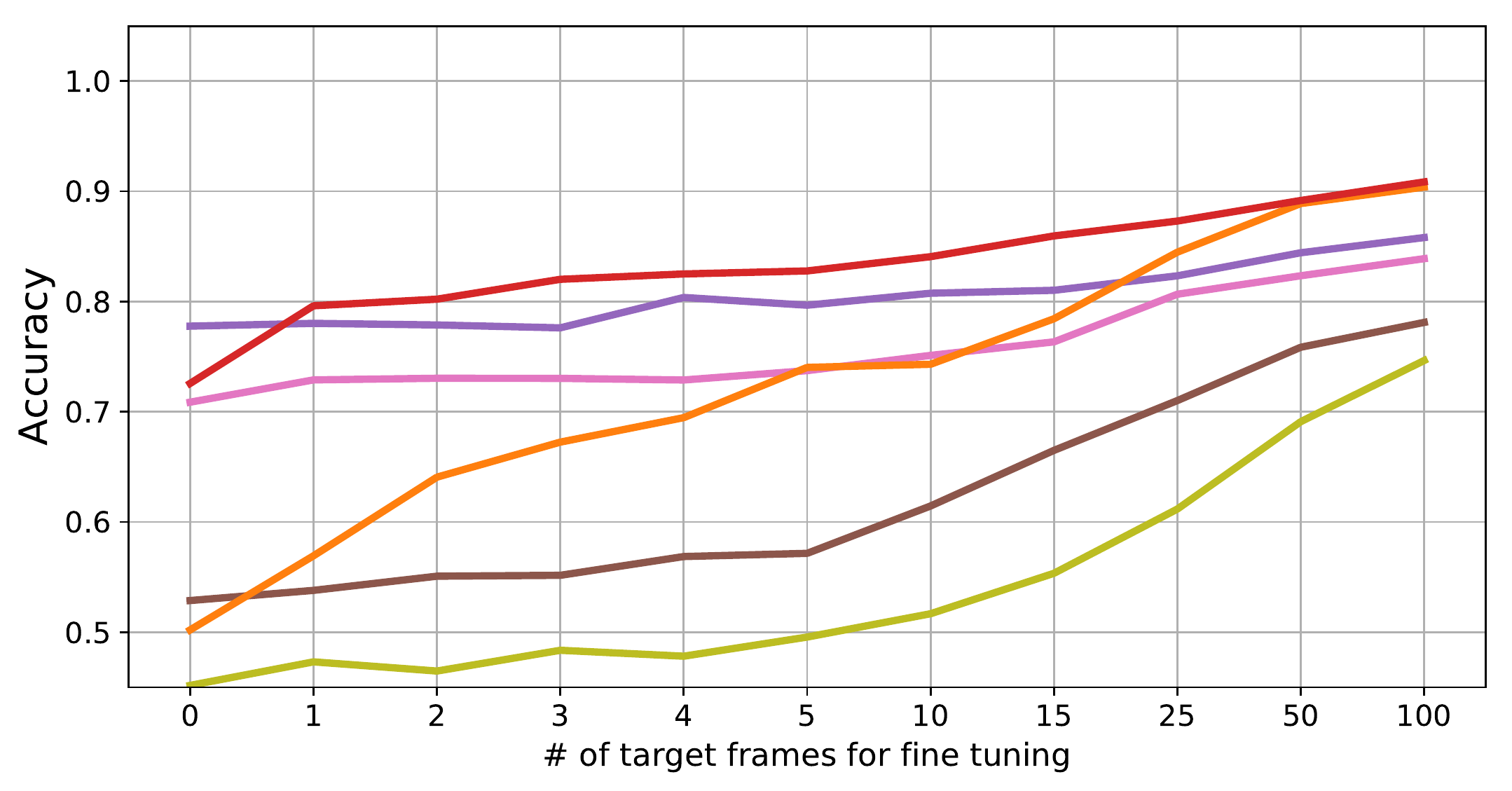} & 
       \includegraphics[width=0.485\linewidth, page=1, trim=0  0 0 -10]{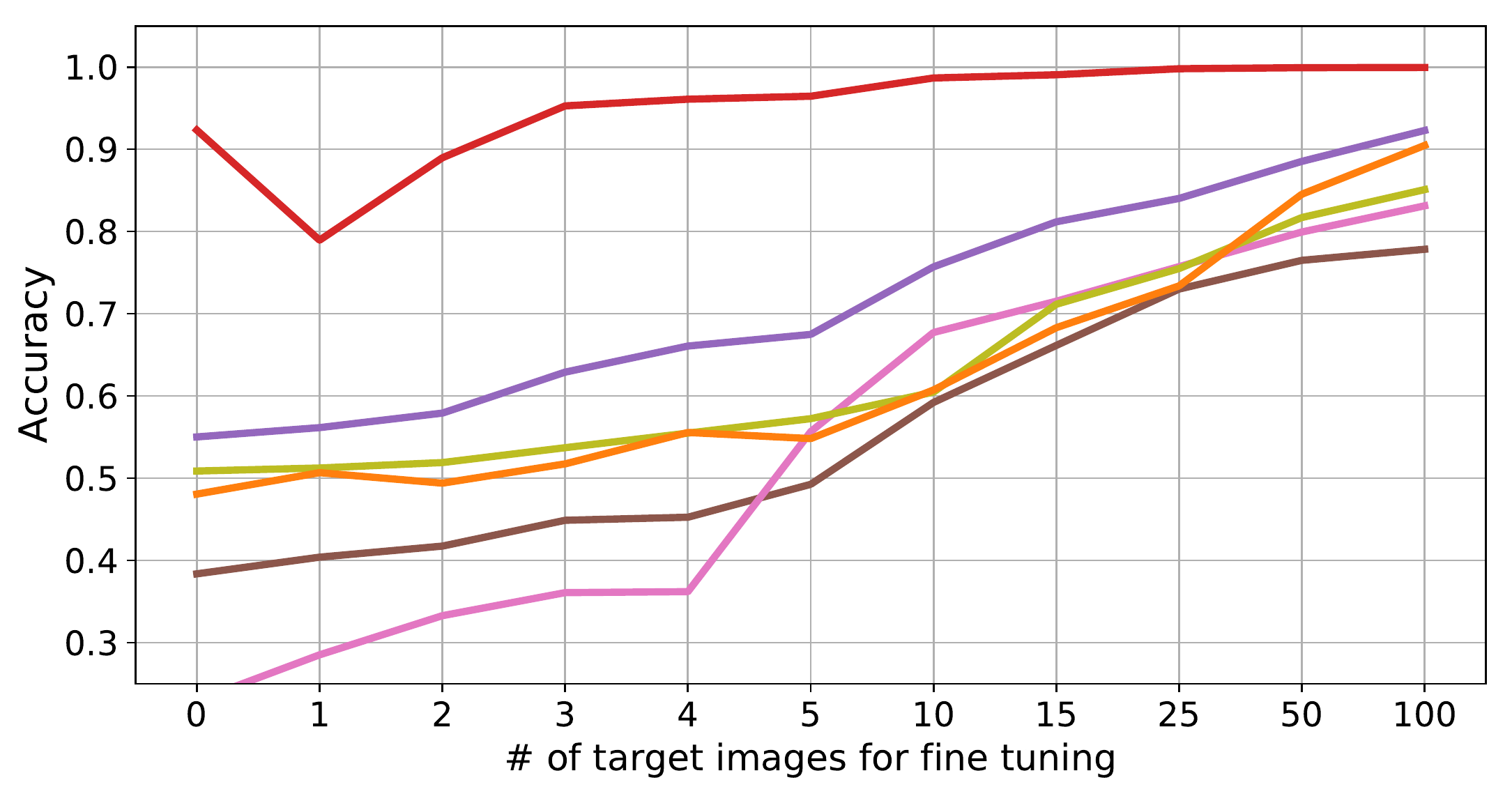}  \\
       {\footnotesize (e) Face2Face \cite{Thies2016face} vs FaceSwap \cite{FaceSwap} }  & 
       {\footnotesize (f)  StarGAN+Glow vs Cycle }  \\
    \end{tabular}
    }
    \vspace{-0.1cm}
    \caption{Few-shot adaptation. Plots show accuracy versus number of images used for fine-tuning (shots). Values are averaged over 10 runs. Note the zero and few-shot transferability of \OURS (FT, red curves).
    }
    \label{fig:ex1comparison}
\end{figure*}

\subsection{Ablation study}
\label{sec:preliminary}
As an ablation study, we compare the performance of the proposed method with four variants obtained by modifying only a specific key aspect.
In the {\em no-residual} variant, we remove the preliminary high-pass filter, thus working on natural images rather than residuals.
In the {\em no-reconstruction} variant we remove the constraint on image reconstruction by setting $\gamma=0$, 
which amounts to canceling the term $\mathcal{L}_{REC}$ from the overall loss.
In the {\em cross-entropy} variant, the proposed $\mathcal{L}_{ACT}$ loss is replaced by a softmax/cross-entropy loss.
Finally, in the {\em large-margin} variant, the proposed $\mathcal{L}_{ACT}$ loss is replaced by a Large-Margin softmax loss \cite{Liu2016}. 
In Fig.\ref{fig:preliminary_exp} we report the accuracy as a function of the number of target samples used for adaptation (shots), 
considering only two datasets pairings for brevity, Inpainting (left) and Pro/Cycle-GAN (right).
In both cases, results fully support our choices. 
During fine-tuning, we also use a small validation set equal to three-fifths of the small training set.
The {\em no-residual} version exhibits a dramatic impairment, with the performance gap closing only after 100 or more examples, 
thus proving the importance of focusing right-away on precious high-pass details.
A similar behavior is observed for the {\em cross-entropy} version and the {\em large-margin} variant, confirming the soundness of the proposed $\mathcal{L}_{ACT}$ loss.
Instead, the reconstruction constraint has only a minor impact on the second pairing, 
but grants a significant improvement on the first one, with a 15\% initial gain which reduces only slowly with new examples.
With both dataset pairings, the proposed methods reaches accuracies above 90\% within five shots.

Fig.~\ref{fig:preliminary_exp_scatter} shows the scatter plots of the two activations, $a_0(x)$ and $a_1(x)$, for a number of test images, both real (green) and fake (red), of the Inpainting pairing.
The first plot on the left shows that the two activations allow a reliable classification of source-domain images.
On target-domain images (second column) the very same classifier fails to detect a good part of the fakes, never seen before.
However, a single-shot adaptation (third column) allows a reasonable separation of the two classes,
which becomes already quite good when using 10 samples, and nearly perfect at 100.
On the second row, we show the same scatter plots when the proposed activation is replaced by the softamx/cross-entropy loss.
Source-domain performance is as good as before, 
but results on the target domain are much worse in comparison to our approach.
While L1 forces the activations to a specific value,  the cross-entropy loss tries to push samples far from the separating hyper-plane irrespectively of the intra-class variance. The low variance gives an advantage when considering the target domain.
As a result, the distribution of the novel manipulation is more easily distinguishable from the concentrated distribution of real images.

\subsection{Comparison with state-of-the-art}
\label{sec:F_comparison}

In this section we are comparing our method to several state-of-the-art CNN-based detection methods.
Bayar16~\cite{Bayar16} and Cozzolino17~\cite{Cozzolino17} have been proposed to detect generic manipulations based on a constrained network.
Rahmouni17~\cite{Rahmouni2017} detects computer-generated images, while MesoInception-4 was proposed \cite{Afchar2018} for deepfake detection.
We are also considering a general-purpose state-of-the-art deep network, XceptionNet \cite{Chollet17}, given the excellent performance shown in \cite{Roessler2018} for facial manipulation detection.
To ensure fairness, all methods are trained exactly on the same dataset as specified in the previous Section.
When necessary, images have been rescaled \cite{Chollet17, Afchar2018} or clipped \cite{Bayar16, Cozzolino17, Rahmouni2017}, to match the size of the network input layer as specified in the original papers.

\vspace{-0.1cm}
\paragraph{Generalization analysis:}

First of all, we evaluate the ability of each network to detect a manipulation performed with the same method seen in the training set (source) and with a different method (target) \textit{without any form of fine-tuning}.
This experiment is important to understand the intrinsic transferability of a network.
We paired the datasets described in Sec.~\ref{sec:datasets} to consider similar types of manipulations.
Results are reported in Tab.~\ref{tab:my_label}.
On the source domain, all methods, including the proposed ones, perform very well, with accuracies close to $100$\%,
which lower only on the video datasets, where frames are subject to a much stronger compression than images of the other datasets.
When evaluating the trained networks on the respective target domain, we observe a dramatic performance loss for all reference methods, most of which provide accuracies close to $50$\%.
Partial exceptions are represented by Cozzolino17, on the inpainting and video datasets, and Rahmouni17, only on videos.
\OURS also shows a performance impairment; however, the residual-based network maintains good performance;
e.g., 90\% accuracy from Cycle~\cite{Zhu2017unpaired} to  Style-GAN~\cite{Karras2019}.

\paragraph{Adaptation analysis:}

In Fig.\ref{fig:ex1comparison}(a-e) we show accuracies as a function of the number of target-domain training samples.
Curves start from zero-shot values as discussed in the generalization paragraph.
For all reference methods, the accuracy grows quite slowly with fine-tuning and in many cases remains largely below $100$\% even at $100$ training images.
\OURS, instead, reaches a very high accuracy with just a few shots, and outperforms all competitors with a large margin.
On Pro/Cycle-GAN (Fig.\ref{fig:ex1comparison}(a)), the accuracy exceeds $90$\% at $4$ shots, and reaches $100$\% before $100$ sample images.
Also on Cycle-GAN/StyleGAN (Fig.\ref{fig:ex1comparison}(b)) results are very good, which shows ForensicTransfer is able to transfer knowledge also from medium-quality to high-quality generation domains.
On StarGAN/Glow (Fig.\ref{fig:ex1comparison}(c)) results are impressive and one shot is already enough.
On Inpainting (Fig.\ref{fig:ex1comparison}(d)), the $90$\% mark is reached with two shots, and almost $100$\% using $100$ training images.
Only for the more challenging Face2Face/Swap experiment (Fig.\ref{fig:ex1comparison}(e)), a slower growth and a smaller gain is observed, 
with XceptionNet almost closing the gap at $100$ shots, both with accuracy exceeding $90$\%.

\subsection{Multi-source experiments}
\label{sec:muli_source}
In the previous experiments we are showing examples of a one-to-one transferability of forgery detection.
Given multiple source domains, we can extend this idea to a many-to-one transfer of knowledge.
To this end, we train the proposed network on a dataset of multiple sources.
Specifically, we use the following methods for our experiment.
As source domain we generate images using StarGAN~\cite{Choi2018StarGAN} and Glow~\cite{kingma2018Glow}.
The target domain images are synthesized by a CycleGAN~\cite{Zhu2017unpaired}.
Based on the images of the two source domains, we train the encoder-decoder network, disentangling the knowledge about real and fake images.
In Tab.~\ref{tab:results_multi} we show the zero-shot results in comparison to a one-to-one transfer.
The multi-source training improves the detection accuracy of unknown forgeries in the target domain by at least $7$\%.
In Fig.~\ref{fig:ex1comparison}(f), we compare the few-shot adaptation w.r.t. the baseline methods.
All methods are trained on both source domains.
As can be seen our approach behaves similarly to the one-to-one experiments.
Our approach outperforms all baselines by more than 25\% when there are less than 5 target domain samples available.

\section{Conclusion}

We introduce \OURS a new method to facilitate transferability between image manipulation domains.
Specifically, we address the shortcomings of convolutional neural networks, that when trained for a given manipulation approach, are unable to achieve good detection results on different edits -- even when they are related.
In comparison to traditional learning techniques, we can achieve significantly higher detection rates in cases where no or only a few training samples are available in the target domain.
Overall, we believe that our method is a stepping stone towards forgery detectors that only need a few training samples of the target domain.

\section{Acknowledgement}
We gratefully acknowledge the support of this research by the AI Foundation, a TUM-IAS Rudolf M\"o{\ss}bauer Fellowship, and Google Faculty Award.
In addition, this material is based on research sponsored by the Air Force Research Laboratory and the Defense Advanced Research Projects Agency under agreement number FA8750-16-2-0204. 
The U.S. Government is authorized to reproduce and distribute reprints for Governmental purposes notwithstanding any copyright notation thereon. The views and conclusions contained herein are those of the authors and should not be interpreted as necessarily representing the official policies or endorsements, either expressed or implied, of the Air Force Research Laboratory and the Defense Advanced Research Projects Agency or the U.S. Government.

\balance
{\small
\bibliographystyle{ieee_fullname}
\bibliography{egbib}

\begin{thebibliography}{10}\itemsep=-1pt

\bibitem{DeepFake}
Deepfake.
\newblock \url{https://www.deepfakes.club/openfaceswap-deepfakes-software/}.

\bibitem{FaceSwap}
Faceswap.
\newblock \url{https://github.com/MarekKowalski/FaceSwap/}.

\bibitem{Afchar2018}
D. Afchar, V. Nozick, J. Yamagishi, and I. Echizen.
\newblock {MesoNet: a compact facial video forgery detection network }.
\newblock In {\em IEEE International Workshop on Information Forensics and
  Security (WIFS)}, 2018.

\bibitem{Agarwal2017JPEGdimples}
S. Agarwal and H. Farid.
\newblock {Photo Forensics from JPEG Dimples}.
\newblock In {\em IEEE International Workshop on Information Forensics and
  Security (WIFS)}, 2017.

\bibitem{elor2017bringingPortraits}
Hadar Averbuch-Elor, Daniel Cohen-Or, Johannes Kopf, and Michael~F. Cohen.
\newblock Bringing portraits to life.
\newblock {\em ACM Transactions on Graphics (Proceeding of SIGGRAPH Asia
  2017)}, 36(4):to appear, 2017.

\bibitem{Bappy2017}
J.H. Bappy, A.K. Roy-Chowdhury, J. Bunk, L. Nataraj, and B.S. Manjunath.
\newblock Exploiting spatial structure for localizing manipulated image
  regions.
\newblock In {\em Computer Vision and Pattern Recognition Workshops}, pages
  4970--4979, 2017.

\bibitem{Bayar16}
B. Bayar and M.C. Stamm.
\newblock A deep learning approach to universal image manipulation detection
  using a new convolutional layer.
\newblock In {\em ACM Workshop on Information Hiding and Multimedia Security},
  pages 5--10, 2016.

\bibitem{Chen2012}
M. Chen, Z. Xu, K.Q. Weinberger, and F. Sha.
\newblock Marginalized denoising autoencoders for domain adaptation.
\newblock In {\em International Conference on Machine Learning}, pages
  1627--1634, 2012.

\bibitem{Chen2019}
W.-Y. Chen, Y.-C. Liu, Z. Kira, Y.-C.~Frank Wang, and J.-B. Huang.
\newblock A closer look at few shot classification.
\newblock In {\em International Conference on Learning Representations}, 2019.

\bibitem{Choi2018StarGAN}
Y. Choi, M. Choi, M. Kim, J.-W. Ha, S. Kim, and J. Choo.
\newblock {StarGAN: Unified Generative Adversarial Networks for Multi-Domain
  Image-to-Image Translation}.
\newblock In {\em IEEE Conference on Computer Vision and Pattern Recognition
  (CVPR)}, 2018.

\bibitem{Chollet17}
F. Chollet.
\newblock {Xception: Deep Learning with Depthwise Separable Convolutions}.
\newblock In {\em IEEE Conference on Computer Vision and Pattern Recognition
  (CVPR)}, 2017.

\bibitem{Cozzolino17}
D. Cozzolino, G. Poggi, and L. Verdoliva.
\newblock Recasting residual-based local descriptors as convolutional neural
  networks: an application to image forgery detection.
\newblock In {\em ACM Workshop on Information Hiding and Multimedia Security},
  pages 1--6, 2017.

\bibitem{Cozzolino2020}
D. Cozzolino and L. Verdoliva.
\newblock {Noiseprint: a CNN-based camera model fingerprint}.
\newblock {\em IEEE Trans. Inf. Forensics Security}, 15:144--159, Jan. 2020.

\bibitem{Csurka2017}
Gabriela Csurka.
\newblock {\em A Comprehensive Survey on Domain Adaptation for Visual
  Applications}, pages 1--35.
\newblock Springer International Publishing, 2017.

\bibitem{Carvalho13:EDI}
T. de Carvalho, C. Riess, E. Angelopoulou, H. Pedrini, and A. Rocha.
\newblock {Exposing digital image forgeries by illumination color
  classification}.
\newblock {\em IEEE Transactions on Information Forensics and Security},
  8(7):1182--1194, 2013.

\bibitem{Farid16}
H. Farid.
\newblock {\em {Photo Forensics}}.
\newblock The MIT Press, 2016.

\bibitem{Ferrara12:IFL}
P. Ferrara, T. Bianchi, A.~De Rosa, and A. Piva.
\newblock {Image Forgery Localization via Fine-Grained Analysis of CFA
  Artifacts}.
\newblock {\em IEEE Transactions on Information Forensics and Security},
  7(5):1566--1577, Oct 2012.

\bibitem{Gidaris2018}
S. Gidaris and N. Komodakis.
\newblock Dynamic few-shot visual learning without forgetting.
\newblock In {\em The IEEE Conference on Computer Vision and Pattern
  Recognition (CVPR)}, June 2018.

\bibitem{Glorot2011}
X. Glorot, A. Bordes, and Y. Bengio.
\newblock Domain adaptation for large-scale sentiment classification: A deep
  learning approach.
\newblock In {\em International Conference on Machine Learning}, pages
  513--520, 2011.

\bibitem{Guera18deepfake}
D. G{\"u}era and E.J. Delp.
\newblock Deepfake video detection using recurrent neural networks.
\newblock In {\em IEEE International Conference on Advanced Video and Signal
  Based Surveillance}, 2018.

\bibitem{hoffman2017cycada}
Judy Hoffman, Eric Tzeng, Taesung Park, Jun-Yan Zhu, Phillip Isola, Kate
  Saenko, Alexei~A Efros, and Trevor Darrell.
\newblock Cycada: Cycle-consistent adversarial domain adaptation.
\newblock {\em arXiv preprint arXiv:1711.03213}, 2017.

\bibitem{Hu2015}
J. {Hu}, J. {Lu}, and Y. {Tan}.
\newblock Deep transfer metric learning.
\newblock In {\em 2015 IEEE Conference on Computer Vision and Pattern
  Recognition (CVPR)}, pages 325--333, June 2015.

\bibitem{Huh2018}
M. Huh, A. Liu, A. Owens, and A.A. Efros.
\newblock Fighting fake news: Image splice detection via learned
  self-consistency.
\newblock In {\em European Conference on Computer Vision (ECCV)}, 2018.

\bibitem{Iizuka2017globally}
Satoshi Iizuka, Edgar Simo-Serra, and Hiroshi Ishikawa.
\newblock {Globally and Locally Consistent Image Completion}.
\newblock {\em ACM Transactions on Graphics (Proc. of SIGGRAPH 2017)},
  36(4):107:1--107:14, 2017.

\bibitem{Karras2018}
Tero Karras, Timo Aila, Samuli Laine, and Jaakko Lehtinen.
\newblock {Progressive Growing of GANs for Improved Quality, Stability, and
  Variation}.
\newblock In {\em International Conference on Learning Representations}, 2018.

\bibitem{Karras2019}
T. Karras et~al.
\newblock A style-based generator architecture for generative adversarial
  networks.
\newblock In {\em CVPR}, 2019.

\bibitem{Kim2018deepvideo}
H. Kim, P. Garrido, A. Tewari, W. Xu, J.~Thies~M. Nie{\ss}ner, P. P{\'e}rez, C.
  Richardt, M. Zollh{\"o}fer, and C. Theobalt.
\newblock {Deep Video Portraits}.
\newblock {\em ACM Transactions on Graphics (TOG)}, 2018.

\bibitem{kingma2018Glow}
D.P. Kingma and P. Dhariwal.
\newblock Glow: Generative flow with invertible 1x1 convolutions.
\newblock {\em arXiv preprint arXiv:1807.03039}, 2018.

\bibitem{Kodirov2017}
E. Kodirov, T. Xiang, and S. Gong.
\newblock Semantic autoencoder for zero-shot learning.
\newblock In {\em IEEE Conference on Computer Vision and Pattern Recognition
  (CVPR)}, pages 4447--4456, 2017.

\bibitem{Liang2018generative}
X. Liang, H. Zhang, L. Lin, and E. Xing.
\newblock Generative semantic manipulation with mask-contrasting gan.
\newblock In {\em European Computer Vision Conference (ECCV)}, 2018.

\bibitem{Liu2016}
W. Liu et~al.
\newblock Large-margin softmax loss for convolutional neural networks.
\newblock In {\em ICML}, 2016.

\bibitem{Liu18}
Y. Liu, Q. Guan, X. Zhao, and Y. Cao.
\newblock Image forgery localization based on multi-scale convolutional neural
  networks.
\newblock In {\em ACM Workshop on Information Hiding and Multimedia Security},
  2018.

\bibitem{Liu2015celabA}
Ziwei Liu, Ping Luo, Xiaogang Wang, and Xiaoou Tang.
\newblock Deep learning face attributes in the wild.
\newblock In {\em IEEE International Conference on Computer Vision (ICCV)},
  pages 3730--3738, Dec 2015.

\bibitem{Lyu14:ERS}
S. Lyu, X. Pan, and X. Zhang.
\newblock Exposing region splicing forgeries with blind local noise estimation.
\newblock {\em International Journal of Computer Vision}, 110(2):202--221,
  2014.

\bibitem{Marra2018}
F. Marra, D. Gragnaniello, D. Cozzolino, and L. Verdoliva.
\newblock {Detection of GAN-Generated Fake Images over Social Networks}.
\newblock In {\em IEEE Conference on Multimedia Information Processing and
  Retrieval}, pages 384--389, 2018.

\bibitem{Motiian2017}
S. Motiian, Q. Jones, S. Iranmanesh, and G. Doretto.
\newblock Few-shot adversarial domain adaptation.
\newblock In {\em Advances in Neural Information Processing Systems (NIPS)},
  pages 6670--6680. 2017.

\bibitem{Pumarola2018GANimation}
A. Pumarola, A. Agudo, A.M. Martinez, A. Sanfeliu, and F. Moreno-Noguer.
\newblock {GANimation: Anatomically-aware Facial Animation from a Single
  Image}.
\newblock In {\em European Conference on Computer Vision (ECCV)}, 2018.

\bibitem{Rahmouni2017}
N. Rahmouni, V. Nozick, J. Yamagishi, and I. Echizeny.
\newblock Distinguishing computer graphics from natural images using
  convolution neural networks.
\newblock In {\em IEEE Workshop on Information Forensics and Security}, pages
  1--6, 2017.

\bibitem{Rao16}
Y. Rao and J. Ni.
\newblock A deep learning approach to detection of splicing and copy-move
  forgeries in images.
\newblock In {\em IEEE International Workshop on Information Forensics and
  Security (WIFS)}, pages 1--6, 2016.

\bibitem{Roessler2018}
A. R{\"{o}}ssler, D. Cozzolino, L. Verdoliva, C. Riess, J. Thies, and M.
  Nie{\ss}ner.
\newblock Faceforensics: A large-scale video dataset for forgery detection in
  human faces.
\newblock {\em arXiv preprint arXiv:1803.09179}, 2018.

\bibitem{ILSVRC15}
Olga Russakovsky, Jia Deng, Hao Su, Jonathan Krause, Sanjeev Satheesh, Sean Ma,
  Zhiheng Huang, Andrej Karpathy, Aditya Khosla, Michael Bernstein,
  Alexander~C. Berg, and Li Fei-Fei.
\newblock {ImageNet Large Scale Visual Recognition Challenge}.
\newblock {\em International Journal of Computer Vision}, 115(3):211--252, Dec
  2015.

\bibitem{Salloum18}
R. Salloum, Y. Ren, and C.~C.~Jay Kuo.
\newblock {Image Splicing Localization using a Multi-task Fully Convolutional
  Network (MFCN)}.
\newblock {\em Journal of Visual Communication and Image Representation},
  51:201--209, 2018.

\bibitem{Snell2017}
J. Snell, K. Swersky, and R. Zemel.
\newblock Prototypical networks for few-shot learning.
\newblock In {\em Advances in Neural Information Processing Systems (NIPS)},
  pages 4077--4087. 2017.

\bibitem{Sung2018}
F. Sung, Y. Yang, L. Zhang, T. Xiang, P.~H.S. Torr, and T.~M. Hospedales.
\newblock Learning to compare: Relation network for few-shot learning.
\newblock In {\em The IEEE Conference on Computer Vision and Pattern
  Recognition (CVPR)}, June 2018.

\bibitem{suwajanakorn2017synthesizing}
Supasorn Suwajanakorn, Steven~M Seitz, and Ira Kemelmacher-Shlizerman.
\newblock {Synthesizing Obama: learning lip sync from audio}.
\newblock {\em ACM Transactions on Graphics (TOG)}, 36(4):95, 2017.

\bibitem{Szegedy2015}
C. Szegedy, Wei Liu, Yangqing Jia, P. Sermanet, S. Reed, D. Anguelov, D. Erhan,
  V. Vanhoucke, and A. Rabinovich.
\newblock Going deeper with convolutions.
\newblock In {\em IEEE Conference on Computer Vision and Pattern Recognition
  (CVPR)}, pages 1--9, June 2015.

\bibitem{Thies2016face}
J. Thies, M. Zollh{\"o}fer, M. Stamminger, C. Theobalt, and M. Nie{\ss}ner.
\newblock {Face2Face: Real-Time Face Capture and Reenactment of RGB Videos}.
\newblock In {\em IEEE Conference on Computer Vision and Pattern Recognition
  (CVPR)}, pages 2387--2395, June 2016.

\bibitem{Tschannen2018}
M. Tschannen, O. Bachem, and M. Lucic.
\newblock Recent advances in autoencoder-based representation learning.
\newblock {\em arXiv preprint arXiv:1812.05069v1}, 2018.

\bibitem{tzeng2017adversarial}
Eric Tzeng, Judy Hoffman, Kate Saenko, and Trevor Darrell.
\newblock Adversarial discriminative domain adaptation.
\newblock In {\em IEEE Conference on Computer Vision and Pattern Recognition
  (CVPR)}, 2017.

\bibitem{Vinyals2016}
O. Vinyals, C. Blundell, T. Lillicrap, K. Kavukcuoglu, and D. Wierstra.
\newblock Matching networks for one shot learning.
\newblock In {\em Advances in Neural Information Processing Systems (NIPS)},
  pages 3630--3638. 2016.

\bibitem{Yerushalmy2011digital}
I. Yerushalmy and H. Hel-Or.
\newblock Digital image forgery detection based on lens and sensor aberration.
\newblock {\em International Journal of Computer Vision}, 92(1):71--91, 2018.

\bibitem{Yu2018generative}
Jiahui Yu, Zhe Lin, Jimei Yang, Xiaohui Shen, Xin Lu, and Thomas~S Huang.
\newblock Generative image inpainting with contextual attention.
\newblock In {\em IEEE Conference on Computer Vision and Pattern Recognition
  (CVPR)}, 2018.

\bibitem{Zhang2019}
X. Zhang, S. Karaman, and S.-F. Chang.
\newblock {Detecting and Simulating Artifacts in GAN Fake Images}.
\newblock In {\em IEEE Workshop on Information Forensics and Security (WIFS)},
  2019.

\bibitem{Zhou2016}
Bolei Zhou, Aditya Khosla, Agata Lapedriza, Aude Oliva, and Antonio Torralba.
\newblock Learning deep features for discriminative localization.
\newblock In {\em IEEE Conference on Computer Vision and Pattern Recognition
  (CVPR)}, pages 2921--2929, 2016.

\bibitem{Zhou18}
P. Zhou, X. Han, V.I. Morariu, and L.S. Davis.
\newblock Learning rich features for image manipulation detection.
\newblock In {\em IEEE Conference on Computer Vision and Pattern Recognition
  (CVPR)}, 2018.

\bibitem{Zhu2017unpaired}
J.Y. Zhu, T. Park, P. Isola, and A.A. Efros.
\newblock {Unpaired image-to-image translation using cycle-consistent
  adversarial networks}.
\newblock In {\em IEEE International Conference on Computer Vision (ICCV)},
  2017.

\bibitem{Zhu2015learning}
J.-Y. Zhu, P. Kr{\"a}henb{\"u}hl, E. Shechtman, and A.A. Efros.
\newblock {Learning a Discriminative Model for the Perception of Realism in
  Composite Images}.
\newblock In {\em IEEE International Conference on Computer Vision (ICCV)},
  2015.

\end{thebibliography}
}

\newpage
\appendix
\section*{Supplemental Material}
Given a database of manipulated images from a source domain, we are able to detect fakes generated with a different manipulation method (target domain).
In this supplemental document, 
we show, through visualization of the class activation maps, the transferability of our proposal respect to the use of a classic classifier (Sec.~\ref{sec:problem}), then 
we report additional ablation studies (Sec.~\ref{sec:ablation}) as well as a comparison to few-shot learning methods (see Sec.~\ref{sec:FS_comparison}).
For reproducibility, we report the hyperparameter of the used baseline methods from the main paper (see Sec.~\ref{sec:baselines}).

\section{Ablation studies}
\label{sec:ablation}

\begin{figure}[b]
    \centering
    \includegraphics[width=0.99\linewidth, page=1]{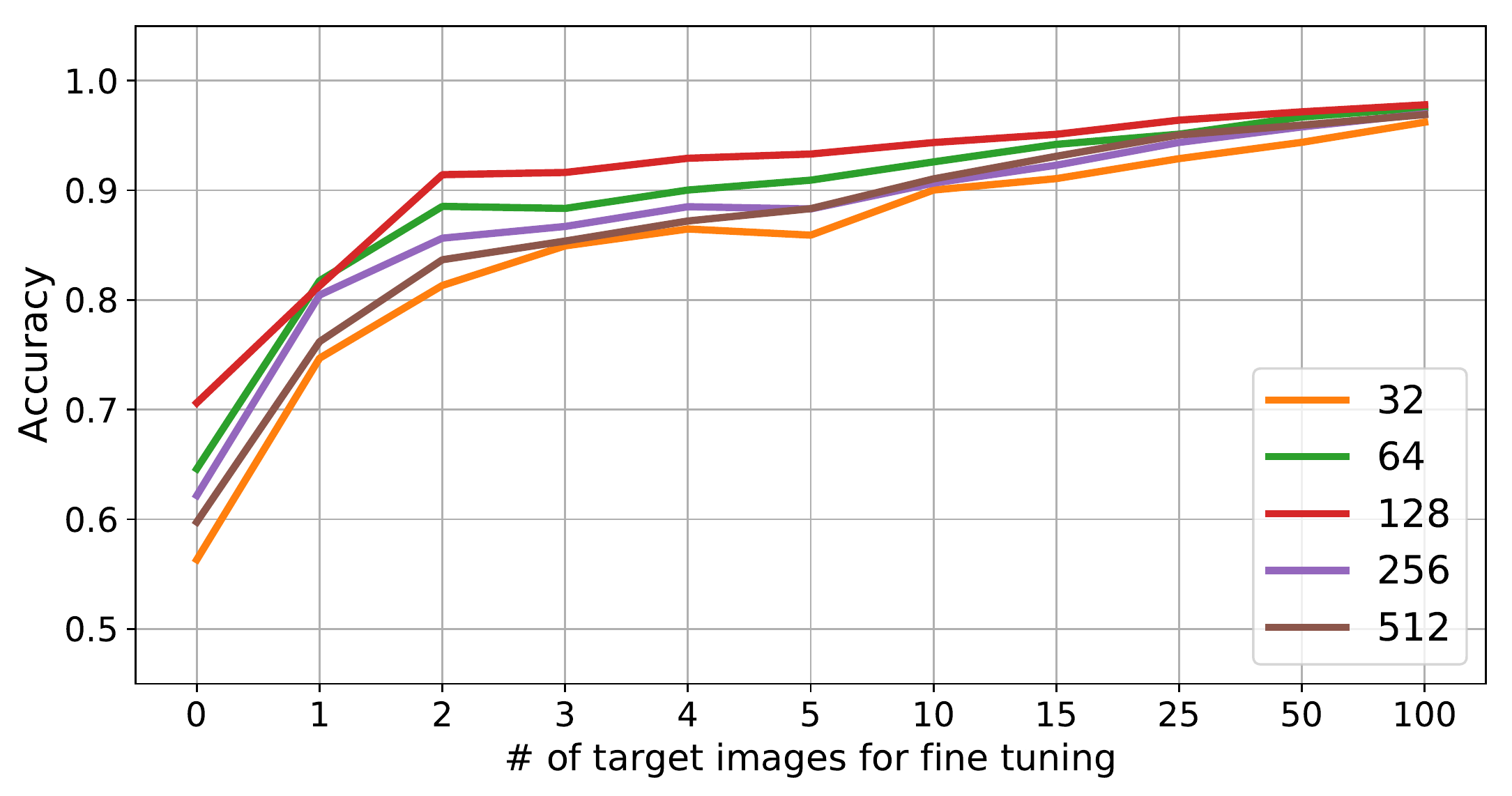}
    \vspace{-0.1cm}
    \caption{
        In this graph, we show the few-shot adaptation abilities with different dimensions of the forensic embedding.
        Plots show accuracy versus number of images used for fine-tuning (shots).
        Values are averaged over 10 runs.
    }
    \label{fig:exScomparison}
\end{figure}

In Fig.~\ref{fig:exScomparison} we show an ablation study with respect to the dimension of the forensic embedding.
The results reported in the main paper are based on a forensic embedding of size $128$, 
where we are using $64$ to encode pristine and another $64$ to encode manipulated images.
In this experiment, we reduce and increase the number of features in the latent space.
Tab.~\ref{tab:dim_graph} shows the zero-shot performance for a one-to-one transfer on the inpainting dataset (Iizuka et al.~\cite{Iizuka2017globally} as source and Yu et al.~\cite{Yu2018generative}  as target domain).
As can be seen the dimensionality has an influence on the performance and $128$ is the sweet spot with the best performance.
With a lower or higher dimensionality of the feature space, the transferability performance decreases,
but it is still better or comparable with respect to the baseline methods.
\newcommand{\ru}{\rule{0mm}{3mm}}
\begin{table}[b!]
    \centering
    \begin{tabular}{l||c|c|c|c|c|} \cline{2-6}
    \ru           &   32  &   64  &   128  &   256  & 512 \\ \hline \hline
    \ru Accuracy  & 56.32 & 64.60 & \bf{70.62} & 62.17 & 59.75 \\ \hline
    \end{tabular}
    \caption{
    Here, we show an ablative study regarding the dimension of the forensic embedding.
    The results are based on a zero-shot transfer scenario on the inpainting dataset.
    }
    \label{tab:dim_graph}
\end{table}

The few-shot adaptation experiment in Fig.~\ref{fig:exScomparison} shows that the performance of the $128$ long feature space is consistently better than the other configurations while they converge with the number of samples from the target domain.

\section{Comparison to few-shot learning methods}
\label{sec:FS_comparison}

Few-shot learning is a relevant problem in many computer vision applications and several methods have been proposed especially with reference to image classification.
In this section, we describe briefly some of such methods and discuss their performance when applied in our scenario.

{\em Chen19} \cite{Chen2019} proposes an approach where a feature extractor is followed by a classifier.
The whole network is trained on the source dataset, 
then the feature extractor is frozen, and only the classifier is fine-tuned on new examples from the target domain.
A ResNet-18 is used for feature extraction, 
while classification is carried out by a cosine similarity layer, which helps reducing intra-class variances \cite{Gidaris2018}.

{\em FADA} \cite{Motiian2017} resorts to adversarial learning to find a feature space 
where the distributions of source and target samples are aligned, ensuring reliable classification in both domains.
As proposed in the reference paper, we use a VGG-16 net for feature extraction, 
a fully connected layer for classification, 
and a network of 2 fully connected layers for the discriminator used in adversarial learning.

{\em MatchingNet} \cite{Vinyals2016}, {\em ProtoNet} \cite{Snell2017}, and {\em RelationNet} \cite{Sung2018}
are similar few-shot learning methods, based on the use of suitable distance metrics.
A network is first used to extract feature vectors, 
and then a suitable distance metric is computed to make predictions, by comparing the input with a small support set of labeled examples.
{\em MatchingNet} uses the cosine similarity, 
{\em ProtoNet} the Euclidean distance with respect to class-mean representation, and 
{\em RelationNet} relies on an appropriately trained CNN.
As feature extractor we use always a ResNet-18, which was proven in \cite{Chen2019} to work well for all these methods.
Then, we train the comparison model in the source domain, and test it using the few examples of the target domain as support set.

When used for few-shot learning on the Inpainting dataset pairing, all these methods exhibit a disappointing performance, 
with accuracy close to 50\% and only slowly growing with the number of examples.
In addition, they are generally much slower than \textit{ForensicTransfer}, especially {\em FADA} due to the adversarial learning.
However, it is worth underlining that all these methods were originally devised for problems like image classification that presents a large number of classes.
Hence, using them in our two-class context may prove sub-optimal.

\begin{figure}[t]
    \centering
    \includegraphics[width=0.99\linewidth, page=1]{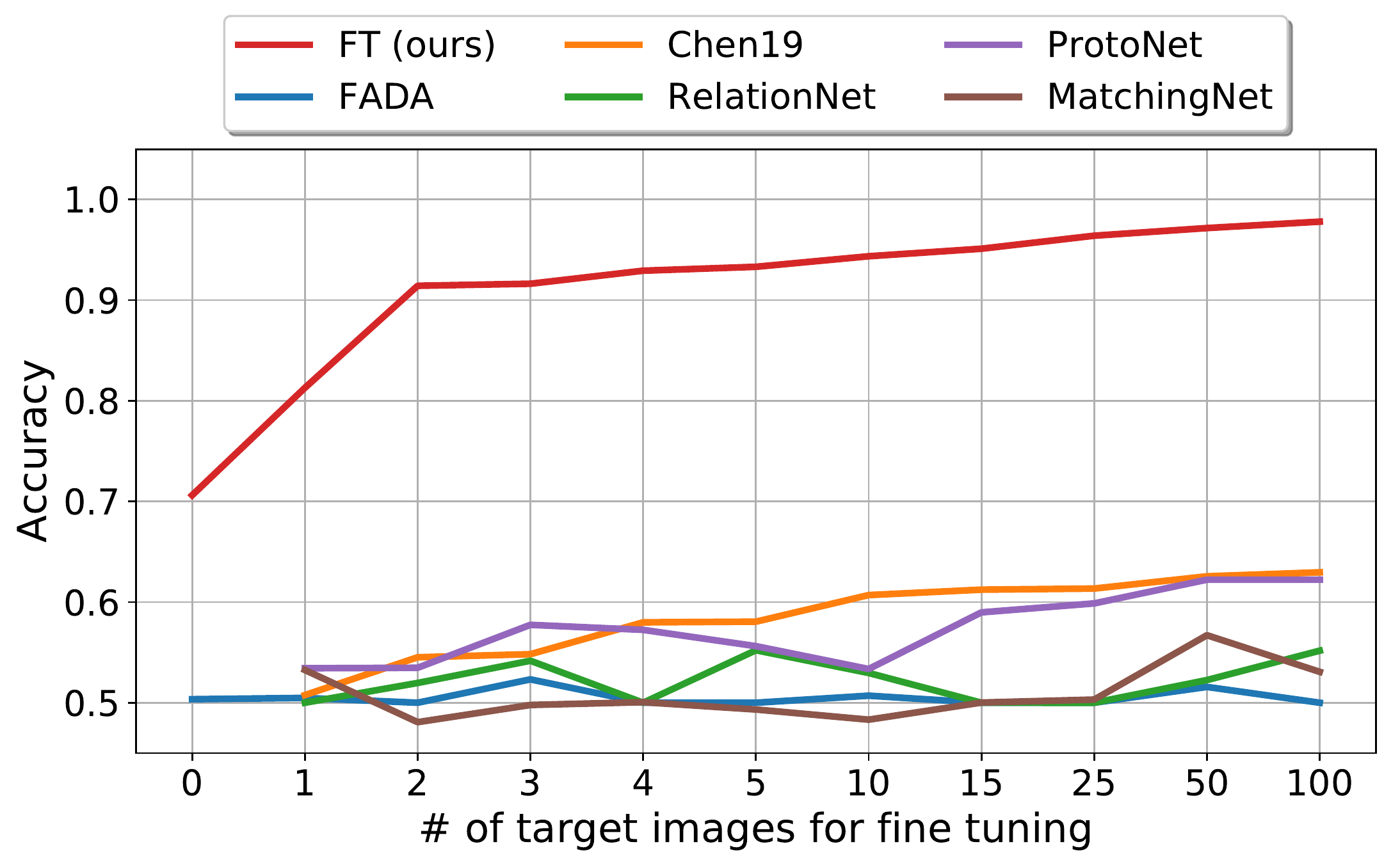}
    \vspace{-0.1cm}
    \caption{
    	Comparison of the proposed method with recent Few-shot learning methods.
        The plot shows accuracy versus number of images used for fine-tuning (shots).
        The results are relative to inpainting manipulations, \cite{Iizuka2017globally} vs \cite{Yu2018generative}.
        The values are averaged over 10 runs.
    }
    \label{fig:ex1fewshot}
\end{figure}

\section{Baseline methods}
\label{sec:baselines}

For the forensic baseline methods used in the main paper, we use the Adam optimizer with the default values for the moments ($\beta_1=0.9$, $\beta_2=0.999$) and different parameters for learning-rate and batch-size.
In the following, we briefly describe these baseline methods:

{\em Bayar16} \cite{Bayar16}: is a constrained network that uses a convolutional layer designed to suppress the high-level content of the image (total of 8 layers). The used learning-rate is $10^{-5}$ with batch-size $64$.
{\em Cozzolino17} \cite{Cozzolino17}: is a CNN-based network that rebuilds hand-crafted features used in the forensic community.
We adopt a learning-rate of $10^{-5}$ and a batch-size of $16$.
{\em Rahmouni17} \cite{Rahmouni2017}: integrates the computation of statistical feature extraction within a CNN framework.
We adopt the best performing network (Stats-2L) with a learning-rate equal to $10^{-4}$ and a batch-size of $64$.
{\em MesoInception-4} \cite{Afchar2018}: is a CNN-based network proposed to detect \textit{DeepFakes}. The network includes inception modules inspired by InceptionNet \cite{Szegedy2015}. 
During training, the mean squared error between true and predicted labels is used as loss instead of the classic cross-entropy loss.
As proposed by the authors, the learning-rate is initially $10^{-3}$ and is reduced by a factor of ten for each $1000$ iterations. The used batch-size is $76$.
{\em XceptionNet} \cite{Chollet17}: is based on depth-wise separable convolution layers with residual connections. XceptionNet is pre-trained on ImageNet.
We adopt a learning-rate of $10^{-3}$ and a batch-size of $64$.
\newline

The training for all methods process is stopped when the validation accuracy does not improve for 10 consecutive epochs and we save the network weights relative to the best validation accuracy.

\begin{figure}
    \centering
    \includegraphics[width=0.99\linewidth]{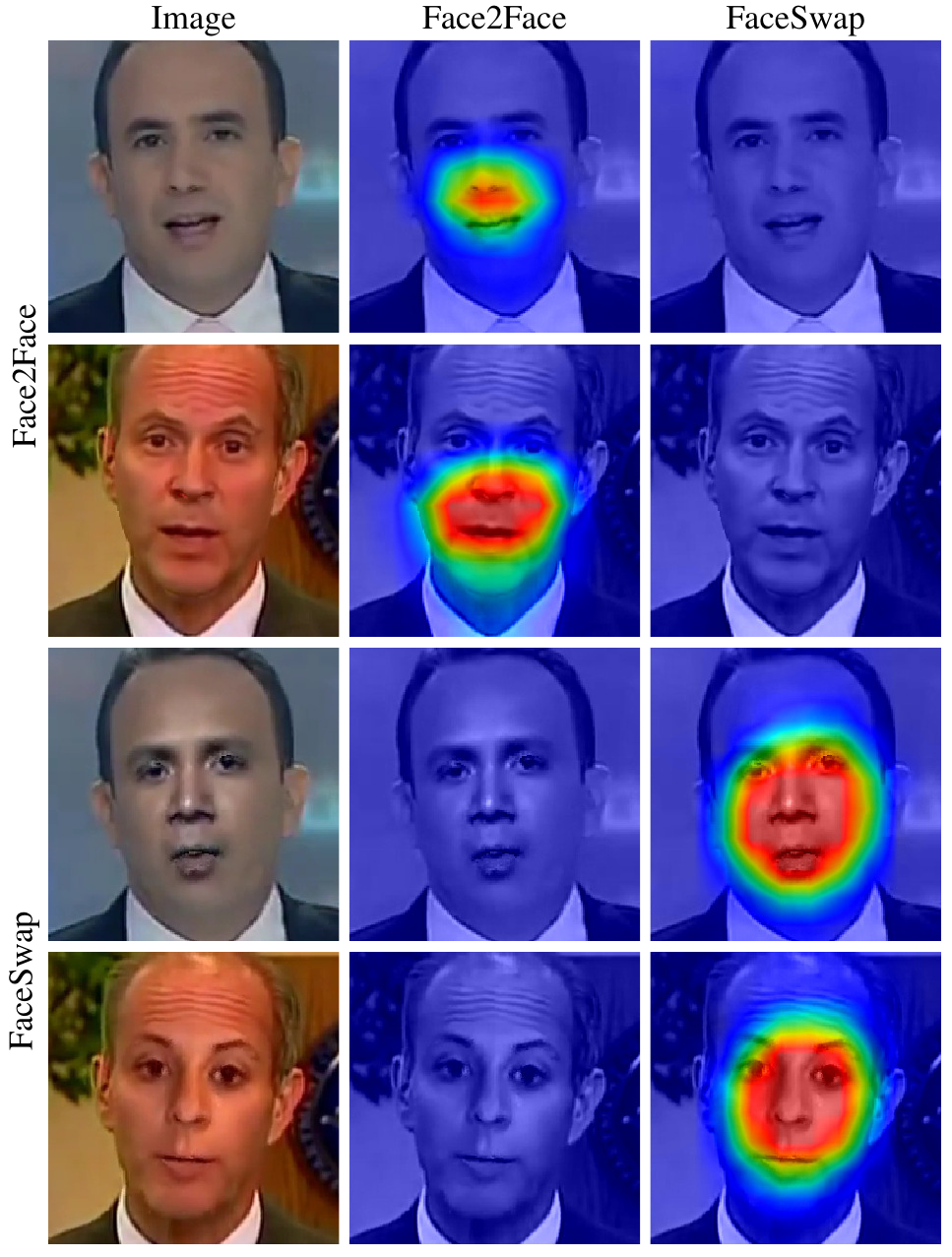}
    \caption{Two examples of images manipulated with Face2Face~\cite{Thies2016face} and FaceSwap~\cite{FaceSwap} (left) and their corresponding class activation maps, when the network (XceptionNet~\cite{Chollet17}) is trained on Face2Face forgeries (middle) and  when it is trained on FaceSwap ones (right).}
    \label{fig:CAM}
\end{figure}

\section{Activation maps}
\label{sec:problem}

In this section we want to carry out a further experiment to gain better insights on the ability of our proposal to transfer to different but closely related manipulations with respect to existing CNN-based detectors.
To showcase this problem, we trained XceptionNet \cite{Chollet17} on images generated by Face2Face \cite{Thies2016face}, which achieves an accuracy of $98.13$\%.
A similar performance is achieved when the network is trained to detect images generated by FaceSwap \cite{FaceSwap}, achieving an accuracy of $98.30$\%.
In Fig.~\ref{fig:CAM}, we show two examples of class activation maps (CAMs), obtained using the approach proposed by \cite{Zhou2016}, for the same subject manipulated with the two approaches.
They clearly show that the network learns to focus on some specific features based on the type of forgery.
As an undesired consequence, a forged image which lacks these artifacts may escape the detector scrutiny.
This becomes problematic when we swap the test sets: now, we train on Face2Face and test on FaceSwap and vice-versa.
Thus, train and test sets are generated from different (even though related) approaches.
Therefore, the cross-domain accuracies drop to 50.20\% and 52.73\%, respectively -- this is only marginally better than random chance.
Our proposal, \OURS, is a CNN-based forensic detector that tackles the problem of detecting novel unseen manipulation methods, without the need of a large amount of training data.
Given a new image from a target domain, one can use the encoder trained on source domain to determine the forensic embedding.
Based on the classification rule mentioned in the main paper, we are able to decide whether the input is closer to the real or to the fake images of the source domain.
If the two manipulation methods share specific artifacts, a fake image of the target domain is closer to the fake images of the source domain.
Given a few samples of the target domain we are able to fine-tune the classifier.

\begin{figure}
    \centering
    \includegraphics[width=0.99\linewidth]{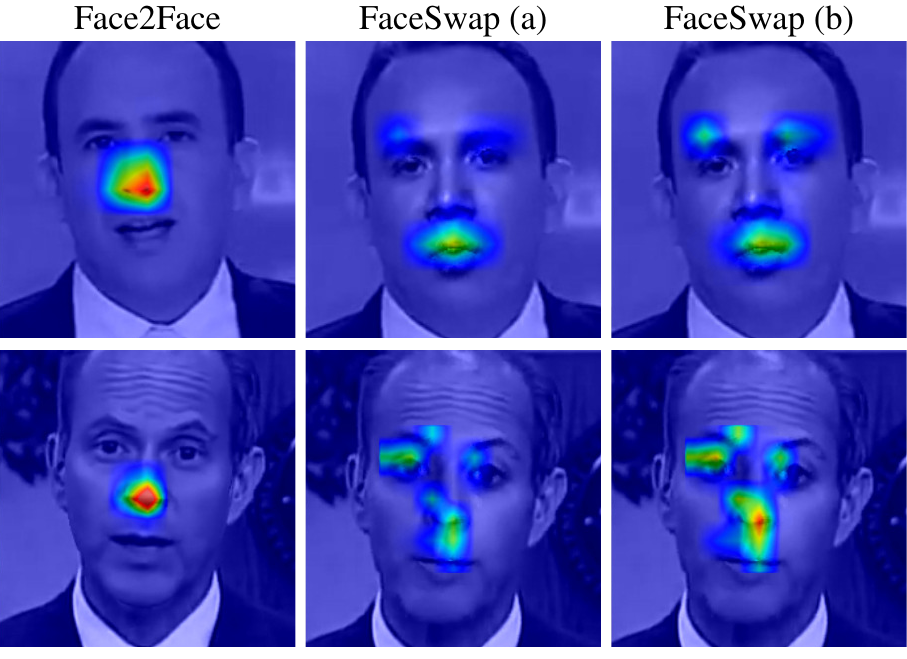}
    \caption{Class activation maps for our method when it is trained on Face2Face~\cite{Thies2016face} and tested on Face2Face forgeries (left) or FaceSwap~\cite{FaceSwap} (middle), and finally trained on Face2Face but fine-tuned using only four images manipulated with FaceSwap and tested on FaceSwap (right).}
    \label{fig:CAMae}
\end{figure}

To proof the transferability of our proposal, we compare the results using our novel approach by visualizing the class activation maps (see Fig.~\ref{fig:CAMae}, cp. to Fig.~\ref{fig:CAM}).
In contrast to a deep CNN like XceptionNet, our approach is able to adapt to the different artifacts of Face2Face and FaceSwap
(the nose in the first case, the eyebrows and mouth in the second case).
Fine-tuning on few images further helps to rely on these different artifacts.

\end{document}